%% file: main.tex
\documentclass{article}

\usepackage[final]{neurips_2018}
\usepackage[utf8]{inputenc} % allow utf-8 input
\usepackage[T1]{fontenc}    % use 8-bit T1 fonts
\usepackage{hyperref}       % hyperlinks
\usepackage{url}            % simple URL typesetting
\usepackage{booktabs}       % professional-quality tables
\usepackage{amsfonts}       % blackboard math symbols
\usepackage{nicefrac}       % compact symbols for 1/2, etc.
\usepackage{microtype}      % microtypography

\usepackage{graphicx}
\usepackage{times}

\usepackage{amssymb,amstext,amsmath, amsthm}

\usepackage{soul,color}
\usepackage{graphicx}
\usepackage{paralist}
\usepackage{caption}
\usepackage{subcaption}
\usepackage{setspace}
\usepackage{multicol}
\usepackage{adjustbox}
\usepackage{array}
\usepackage{multicol}

\usepackage{algorithm}
\usepackage{algorithmic}

\usepackage{todonotes}

\usepackage{wrapfig}

\long\def\remark#1{% for notes in the margin -- from Norman Ramsey
    \ifvmode\else
        \unskip\raisebox{-3.5pt}[0pt][0pt]{\rlap{$\scriptstyle\diamond$}}%
    \fi
    \setlength\marginparwidth{1.5cm}
    \marginpar{\raggedright\hbadness=10000
    \parindent=8pt \parskip=2pt
    \def\baselinestretch{0.8}\tiny
    \itshape\noindent #1\par}}

\newcolumntype{V}{>{\centering\arraybackslash} m{.4\linewidth} }

\input{preamble}

\usepackage{titlesec}
\titlespacing\section{0pt}{4pt plus 2pt minus 2pt}{-2pt plus 2pt minus 0pt}
\titlespacing\subsection{0pt}{4pt plus 2pt minus 2pt}{-2pt plus 2pt minus 0pt}
\titlespacing\subsubsection{0pt}{4pt plus 2pt minus 2pt}{-2pt plus 2pt minus 0pt}

\title{Faithful Inversion of Generative Models\\ for Effective Amortized Inference}

\author{
  Stefan Webb\thanks{Correspondence to \texttt{info@stefanwebb.me}} \\
  University of Oxford \\
  \And
  Adam Goli\'{n}ski \\
  University of Oxford \\
  \And
  Robert Zinkov \\
  UBC \\
  \AND
  N. Siddharth \\
  University of Oxford \\
  \And
  Tom Rainforth \\
  University of Oxford \\
  \And
  Yee Whye Teh \\
  University of Oxford \\
  \And
  Frank Wood \\
  UBC \\
}

\begin{document}
\suppressfloats

\setlength{\abovedisplayskip}{3.5pt}
\setlength{\belowdisplayskip}{3.5pt}
\setlength{\abovedisplayshortskip}{3.5pt}
\setlength{\belowdisplayshortskip}{3.5pt}	

\maketitle

\begin{abstract}
  \input{abstract}
\end{abstract}

\input{introduction}

\input{method}

\input{applications}

\input{discussion}

\vfill\pagebreak
\input{acknowledgements}

\bibliography{longstrings,dphil}
\bibliographystyle{icml2018}

\end{document}

\suppressfloats

\maketitle

\section{Probabilistic Graphical Models}
\label{sec:pgm-overflow}

This summary is based on \citet{KollerFriedman2009}.

\subsection{Bayesian networks and representation}\label{sec:bn-representation}
Any probability distribution implicitly represents certain independence relationships between its variables via its factorization.
These are of interest because they can be exploited to both compactly represent distributions and to reduce the cost of inference.
The set of such relationships is defined as:
\begin{definition}
	Let $p$ be a distribution defined over $\mathcal{X}$.
We define $\mathcal{I}(p)$ to be the set of \emph{independence assertions} of the form $(\mathbf{X}\perp\mathbf{Y}\mid\mathbf{Z})$ that hold in $p$, where $\mathbf{X},\mathbf{Y},\mathbf{Z}\subseteq\mathcal{X}$.
\end{definition}
The framework of probabilistic graphical models is used for representing and reasoning about a wide class of probability distributions by making these independence assertions explicit.
Distributions are represented as the product of factors over subsets of the model variables.
Associated with the factorization is a graph, wherein the nodes are the random variables of the model, and the edges express the distribution's independence assertions.

Bayesian networks (BNs) are a class of probabilistic graphical models that use a directed acyclic graph.
We refer to the graph alone as the BN structure, whereas the BN itself comprises, in addition, a representation for each factor.
In a BN, each variable has a conditional distribution that only depends on its parents in the graph.
For example, in Figure \ref{fig:example1-bn} the distribution factors as $p(a)p(b|a)p(c|a)p(d|b)p(e|c)$.

Formally, the semantics of the BN structure are that it encodes the local independencies:
\begin{definition}
	A Bayesian network structure $\mathcal{G}$ encodes the \emph{local independencies} $\mathcal{I}_l(\mathcal{G})$, namely, those of the form $X_i \perp \textnormal{NonDescendants}_{X_i}\mid\textnormal{Pa}^\mathcal{G}_{X_i}$ for each $X_i\in\mathcal{G}$, where $\textnormal{Pa}^\mathcal{G}_{X_i}$ denotes the parents of $X_i$ in $\mathcal{G}$.
\end{definition}
It turns out that there are additional independencies that can be read off $\mathcal{G}$ aside from the local ones, that hold for every $p$ that factorizes over $\mathcal{G}$, and these are identified by the concept of \emph{d-separation}.

We relate the conditional independencies encoded in a graph, such as a BN structure, to a corresponding distribution by the concept of an independency map, or \emph{I-map}:
\begin{definition}
	Let $\mathcal{K}$ be any graph object associated with a set of independencies $\mathcal{I}(\mathcal{K})$.
We say that $\mathcal{K}$ is an \emph{I-map} for a distribution $p$ if $\mathcal{I}(\mathcal{K})\subseteq\mathcal{I}(p)$.
\end{definition}
In our case, a BN structure $\mathcal{G}$ is an I-map for $p$ if $\mathcal{I}_l(\mathcal{G})\subseteq\mathcal{I}(p)$.
This means that $\mathcal{G}$ may not encode all the independencies in $p$, \emph{but it does not mislead us by encoding independencies not present in $p$}.
For this reason, we will interchangeability use the expression, ``$\mathcal{G}$ is faithful to $p$.''

It can be proven that a BN structure $\mathcal{G}$ is an I-map for a distribution $p$ if and only if $p$ is representable as a set of conditional probability distributions (also referred to as model factors), factoring according to $\mathcal{G}$, that is,
\begin{align*}
	P(\mathcal{X}) &= \prod_{X_i\in\mathcal{X}}P(X_i\mid\textnormal{Pa}^\mathcal{G}_{X_i}).
\end{align*}
Therefore, we can use the graph as a means of revealing the structure in a distribution.

\subsection{D-separation}
\label{sec:d-separation}
We give a heuristic explanation of d-separation by examining the opposite question of, roughly speaking, when can probabilistic influence flow from one variable to another.

In paths in $\mathcal{G}$ with three variables that form,
\begin{itemize}
	\item a causal trail, $X\rightarrow  Z\rightarrow Y$,
	\item an evidential trail, $X\leftarrow Z\leftarrow Y$, or,
	\item a common cause, $X\leftarrow Z\rightarrow Y$,
\end{itemize}
knowledge of $X$ is informative about $Y$ when $Z$ is not observed, and observing $Z$ blocks this flow of information.
For example, suppose $X$ is the coherence of a course, $Z$ its difficulty, and $Y$ the grade a student receives.
Further, suppose there is a causal trail $X\rightarrow  Z\rightarrow Y$ in the graph and no other trail between $X$ and $Y$.
If we observe that the course is taught coherently, this will inform our beliefs about its difficulty, which will in turn change our beliefs about the student's grade.
On the other hand, if we observe that it is a difficult course, the coherency of the course will not effect our beliefs about the student's grade as it can only do so indirectly via the difficulty variable.

Conversely, for a common effect motif, $X\rightarrow Z\leftarrow Y$, also known as a \emph{v-structure}, there is an ``explaining away'' effect, whereby if we observe $Z$ (or a descendent of $Z$), then knowledge of $X$ \emph{is} informative about $Y$.
For example, if $X$ if the difficulty of an exam, $Z$ is a student's result, and $Y$ is his aptitude, then if we observe a poor result and that the exam is hard, we can attribute the result to the difficulty of the exam, and lessen our belief that the student is incapable.

This heuristic reasoning generalizes to longer trails in the concept of an \emph{active trail},
\begin{definition}
Let $\mathcal{G}$ be a BN structure and $X_1\rightleftharpoons\cdots\rightleftharpoons X_n$ a trail in $\mathcal{G}$.
Let $\mathbf{Z}$ be a subset of observed variables.
The trail $X_1\rightleftharpoons\cdots\rightleftharpoons X_n$ is \emph{active} given $\mathbf{Z}$ if,
\begin{itemize}
	\item Whenever we have a v-structure $X_{i-1}\rightarrow X_i\leftarrow X_{i+1}$, then $X_i$ or one of its descendants are in $\mathbf{Z}$;
	\item No other node along the trail is in $\mathbf{Z}$.
\end{itemize}
\end{definition}
Those subsets of variables, conditioned on another set, are said to be d-separated if an active trail does not exist between them.
Formally:
\begin{definition}
	Let $\mathbf{X}, \mathbf{Y}, \mathbf{Z}$ be three sets of nodes in $\mathcal{G}$.

We say that $\mathbf{X}$ and $\mathbf{Y}$ are d-separated given $\mathbf{Z}$, denoted $\textnormal{d-sep}_\mathcal{G}(\mathbf{X};\mathbf{Y}\mid\mathbf{Z})$, if there is no active trail between any node $X\in\mathbf{X}$ and $Y\in\mathbf{Y}$ given $\mathbf{Z}$.
We use $\mathcal{I}(\mathcal{G})$ to denote the set of independencies that correspond to d-separation,
	\begin{align*}
		\mathcal{I}(\mathcal{G}) &= \{\left(\mathbf{X}\perp\mathbf{Y}\mid\mathbf{Z}\right)\mid \textnormal{d-sep}_\mathcal{G}\left(\mathbf{X};\mathbf{Y}\mid\mathbf{Z}\right)\}.
	\end{align*}
\end{definition}

D-separation is sound in the sense that if $\mathbf{X}$ and $\mathbf{Y}$ are d-separated given $\mathbf{Z}$ in a graph $\mathcal{G}$, then $\mathbf{X}\perp\mathbf{Y}\mid\mathbf{Z}$ holds in all distributions $p$ that factorize according to $\mathcal{G}$ \citep[Theorem 3.3]{KollerFriedman2009}.

A certain converse statement also holds for the completeness of d-separation.
If $\mathbf{X}$ and $\mathbf{Y}$ are not d-separated given $\mathbf{Z}$ in a graph $\mathcal{G}$, then $\mathbf{X}\perp\mathbf{Y}\mid\mathbf{Z}$ does not hold for almost all (in a measure theoretic sense) distributions $p$ that factorize according to $\mathcal{G}$ \citep[Theorem 3.5]{KollerFriedman2009}.
So, for all practical purposes one may assume $\mathcal{I}(\mathcal{G})=\mathcal{I}(p)$.

\subsection{Exact inference by variable elimination}
\label{sec:variable-elimination}
Variable elimination is an algorithm for performing exact inference in graphical models which have the property that summation of variables in the model factors is tractable---typically ones with discrete finite-valued factors.
From a higher perspective, it works by using the observation that we can exchange the order of the summation of the model variables and the multiplication of the model factors based on their scope, i.e.\ what variables they take as inputs.
Doing so can greatly reduce the complexity of summation, or rather inference, if the variable ordering is carefully chosen.

Consider the BN structure from Figure \ref{fig:student-bn-plain} and suppose the task is to compute $P(J)$.
Simply multiplying all the factors together, then summing out $\mathcal{X}\setminus\{J\}$,
\begin{align*}
	P(J) &= \sum_{\mathcal{X}\setminus\{J\}}\prod_{X\in\mathcal{X}}\phi_X,
\end{align*}
would not be an efficient means to do so.
Rather, we ought to exploit the structure in the model, and perform summation on factors with smaller scope.
Suppose also, that we perform the summation, or variable elimination, in the ordering $D,I,H,G,S,L$.
To sum out $D$, we can pull out all factors that do not contain $D$ in their scope.
First we multiply the factors depending on $D$ together,
\begin{align*}
	\psi_1(D) &= \phi_D(D)\phi_G(G,I,D),
\end{align*}
then sum out $D$,
\begin{align*}
	\tau_1(G,I) &= \sum_D\psi_1,
\end{align*}
to produce a new intermediate factor that is used in subsequent computations.

Similarly, to sum out $I$,
\begin{align*}
	\psi_2(G,I,S) &= \tau_1(G,I)\phi_I(I)\phi_S(S,I),\\
	\tau_2(G,I) &= \sum_I\psi_2.
\end{align*}
continuing this process to eliminate the remaining variables.
As each intermediate factor, $\psi_i$, has a scope much narrower than the full variables set, $\mathcal{X}$, exact inference is made tractable.

 \subsection{Induced graphs}
 \label{sec:induced-graphs}

\begin{figure*}[t]
    \centering
		\begin{subfigure}[t]{0.15\textwidth}
				\centering
        \includegraphics[width=0.9\textwidth]{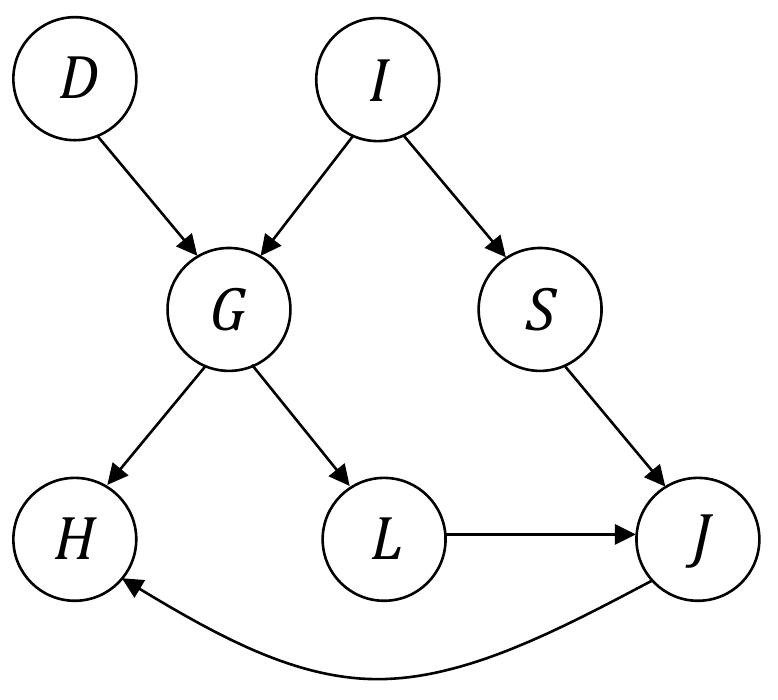}
        \caption{}
        \label{fig:student-bn-plain}
    \end{subfigure}\hfill
		\begin{subfigure}[t]{0.15\textwidth}
				\centering
        \includegraphics[width=0.9\textwidth]{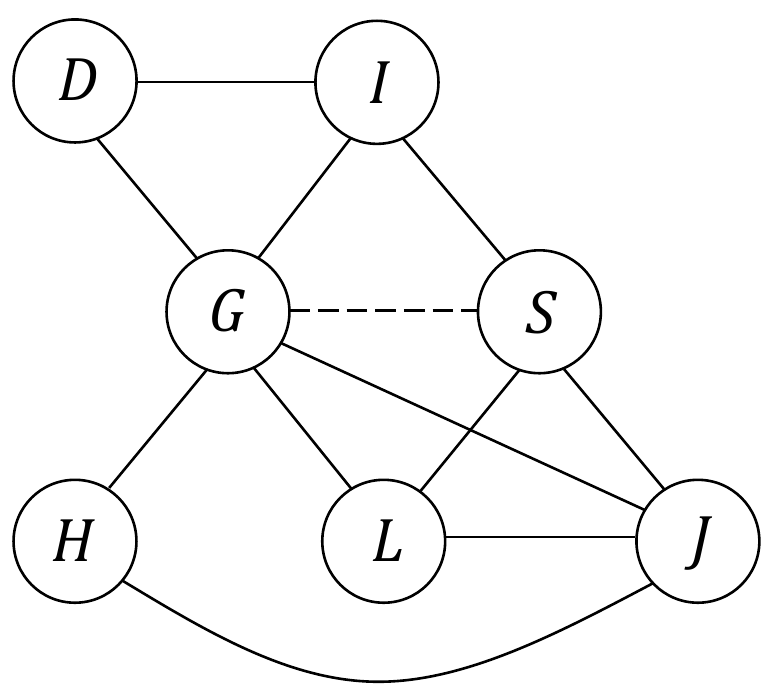}
        \caption{}
        \label{fig:student-induced-2}
    \end{subfigure}\hfill
    \begin{subfigure}[t]{0.35\textwidth}
				\centering
        \includegraphics[width=1.0\textwidth]{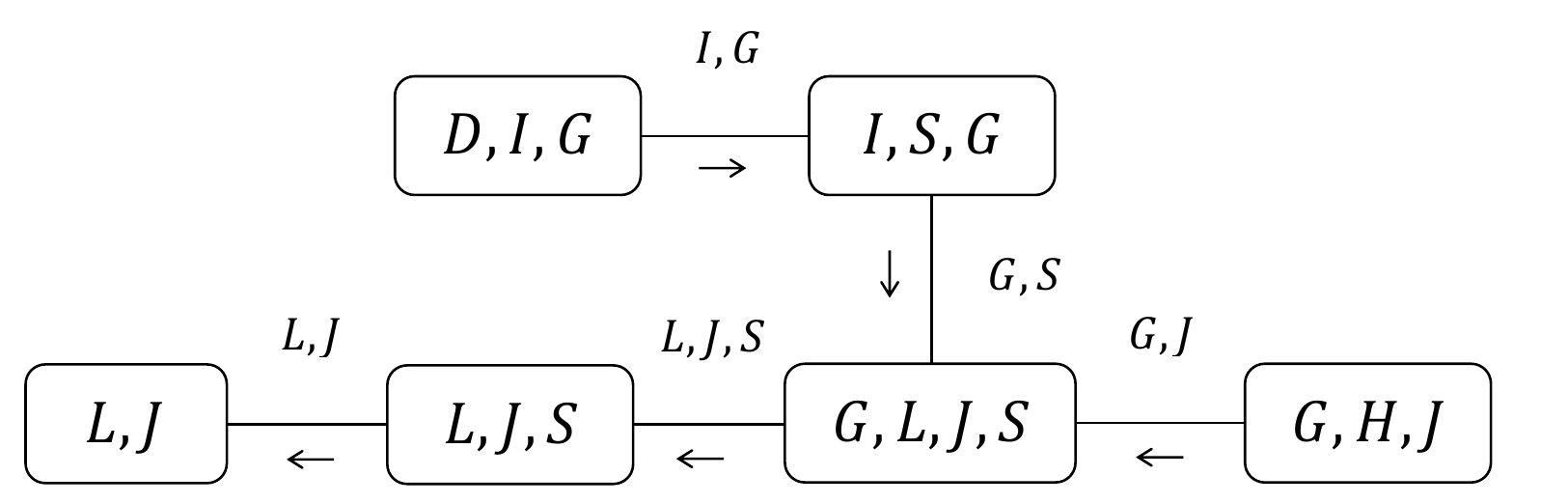}
        \caption{}
        \label{fig:student-clique-tree-2}
    \end{subfigure}\hfill
		\begin{subfigure}[t]{0.35\textwidth}
				\centering
        \includegraphics[width=1.0\textwidth]{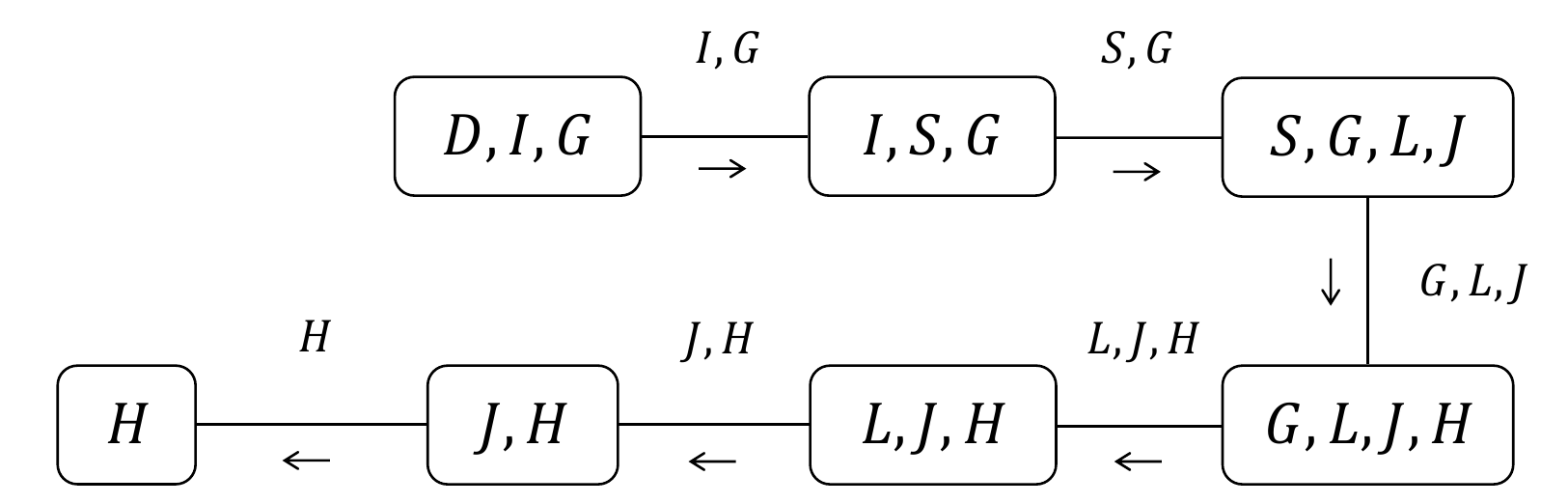}
        \caption{}
        \label{fig:student-clique-tree}
    \end{subfigure}
    \caption[``Extended Student'' example]{(a) BN structure for ``Extended Student'' example; (b) the induced graph corresponding to elimination ordering $D,I,H,G,S,L$; (c) the corresponding clique tree; (d) the clique tree corresponding to elimination ordering $D,I,S,G,L,J,H$.}
    \label{fig:student-graphs}
\end{figure*}

 The computational cost of an application of variable elimination, which depends on the size of the scope of the largest intermediate factor, can be captured in an undirected graph known as the \emph{induced graph}.
 It is defined as follows:
 \begin{definition}
 	Let $\Phi$ be a set of factors over $\mathcal{X}=\{X_1,\ldots,X_n\}$, and $\prec$ be an elimination ordering for some subset $\mathcal{X}\subseteq\mathcal{X}$.
  The induced graph $\mathcal{I}_{\Phi,\prec}$ is an undirected graph over $\mathcal{X}$, where $X_i$ and $X_j$ are connected by an edge if they both appear in some intermediate factor $\phi$ generated by the variable elimination algorithm using $\prec$ as an elimination ordering.
 \end{definition}
 The induced graph for our previous example is given in Figure \ref{fig:student-induced-2}.
 We see that it has cliques, or maximally connected subgraphs, for the subsets  $\{D,I,G\}$, $\{I,S,G\}$, $\{G,J,S,L\}$, and $\{G,H,J\}$, which correspond to the scopes of some intermediate factor, $\psi_i$, in the computation.

  We can form the induced graph for a given run of variable elimination on $\mathcal{G}$ as follows.
  First, we ``moralize'' $\mathcal{G}$ by connecting all its parents and removing the directionality of the edges.
  This induces an edge between $X_i$ and $X_j$ if they appear in the scope of a model factor $\phi\in\Phi$ before variable elimination.
  During variable elimination, after we have calculated the scope of each intermediate factor, we add additional edges to the graph, indicated in our figures with dotted edges, so that the scope of each intermediate factor, $\psi_i$, is maximally connected.
  For instance, in our example, when  eliminating $I$, a factor $\psi_3(G,I,S)$ occurs, so we must add the additional edge $G-S$.
  A good variable elimination ordering will add as few additional edges so that the scope of the intermediate factors is constrained.

\subsection{Clique trees}\label{sec:clique-trees}

Another way to understand the variable elimination algorithm is as an algorithm that passes messages over a tree structure known as a clique tree.
Continuing our running ``Student'' example, the clique tree corresponding to the variable elimination ordering $D,I,H,G,S,L$ is given in Figure \ref{fig:student-clique-tree-2}.
We refer to the nodes in the tree as the cliques, which are subsets of the model variables corresponding to the scopes of the intermediate factors, $\{\psi_i\}$.
Each model factor, $\phi_i$, is associated to a node in the graph, for example, $\phi_D(D)$, $\phi_G(D,I,G)$, and $\phi_I(I)$ are associated with the node ``$D,I,G$,'' and $\phi_S(I,S)$ is associated with ``$I,S,G$.''

The messages, $\{\tau_i\}$, are formed by multiplying together all the factors associated with a node and its incoming messages, and summing out the variables not in the intersection of the node and its downstream neighbour.
The intersections of the node scopes are indicated above each edge and are known as the sepsets.
The tree is undirected, although we have indicated the directionality of message passing with arrows above each edge.

Formally, a clique tree is defined as follows:
\begin{definition}
	A clique tree $\mathcal{U}$ for a set of factors $\Phi$ over $\mathcal{X}$ is an undirected graph, each of whose nodes $i$ is associated with a subset $\mathbf{C}_i\subset\mathcal{X}$.
A clique tree must be family-preserving---each factor $\phi\in\Phi$ must be associated with a clique $\mathbf{C}_i$ such that $\textnormal{scope}[\phi]\subseteq\mathbf{C}_i$.
Each edge between a pair of cliques $\mathbf{C}_i$ and $\mathbf{C}_j$ is associated with a \emph{sepset} $\mathbf{S}_{i,j}\subseteq \mathbf{C}_i\cap\mathbf{C}_j$.
Also, it must hold that whenever there is a variable $X$ such that $X\in\mathbf{C}_i$ and $X\in\mathbf{C}_j$, then $X$ is also in every clique in the (unique) path in $\mathcal{T}$ between $\mathbf{C}_i$ and $\mathbf{C}_j$.
\end{definition}

An important property of clique trees, known as the \emph{sepset property}, is the following: all variables upstream of a clique are conditionally independent of those downstream, conditioned on the corresponding sepset, and the sepset is the minimal set for which this holds \citep[Theorem 10.2]{KollerFriedman2009}.
In this way, the sepset ``separates'' upstream and downstream variables.
Property 1 in B.4 is equivalent to the sepset property---our definition of ``upstream/downstream'' coincides in induced graphs and clique trees, and the sepsets are seen to correspond to the downstream neighbours of a variable.
Compare the induced graph of \S2.2 with its corresponding clique tree in Figure \ref{fig:student-clique-tree}.

\subsection{Exact inverses}
Is it possible in general for a stochastic inverse $\mathcal{H}$ to perfectly capture the independencies in $\mathcal{G}$ so that $\mathcal{I}(\mathcal{H})=\mathcal{I}(\mathcal{G})$? The answer is given in the negative by the following theorem and associated definitions \citep[Theorem 3.8]{KollerFriedman2009}:
\begin{definition}
	The \emph{skeleton} of a BN structure $\mathcal{G}$ over $\mathcal{X}$ is an undirected graph over $\mathcal{X}$ that contains an edge $\{X,Y\}$ for every edge $(X,Y)$ in $\mathcal{G}$.
\end{definition}
\begin{definition}
	A v-structure $X\rightarrow Y\leftarrow Z$ is an \emph{immorality} if there is no direct edge between $X$ and $Y$.
\end{definition}
\begin{theorem}
	Let $\mathcal{G}$ and $\mathcal{H}$ be two graphs over $\mathcal{X}$.
Then $\mathcal{G}$ and $\mathcal{H}$ have the same skeleton and the same set of immoralities if and only $\mathcal{I}(\mathcal{H})=\mathcal{I}(\mathcal{G})$.
\end{theorem}
In general, immoralities in $\mathcal{G}$ are destroyed in $\mathcal{H}$, as both heuristic and faithful inversion methods may reverse edges in v-structures or add a direct edge between their parents.

\section{Restrictions on orderings}
So far, we have been simulating variable elimination on the latent variables in the model, stopping at the observed ones. In special cases, we may wish to further restrict the variable elimination ordering within the non-observed variables. For instance, the semi-supervised variational objective of \citet{Kingma2014Semi} requires a factorization $q(\mathbf{z},\mathbf{y}\mid\mathbf{x})=q(\mathbf{z}\mid\mathbf{x},\mathbf{y})q(\mathbf{y}\mid\mathbf{x})$, where $\mathbf{y}$ are the semi-observed variables. In this case we should eliminate all $\mathbf{z}$ before eliminating $\mathbf{y}$. Algorithm 1 can be suitably modified to accommodate this by running Lines 6--17, replacing ``latents'' and ``latent variables'' with $z\in\mathbf{z}$, and repeating Lines 6--16 replacing those terms with $y\in\mathbf{y}$. In a time series model, we may wish to eliminate the latent variables in their time ordering, $\mathbf{z_1},\ldots,\mathbf{z_T}$, and can repeat Lines 6--16 $T$ times, replacing those terms with $z\in\mathbf{z_i}$ in turn.

\section{Counterexamples to Stuhlm\"{u}ller's heuristic inversion}
\begin{figure*}[t]
    \centering
		\begin{subfigure}[b]{0.16\textwidth}
				\centering
        \includegraphics[scale=0.32]{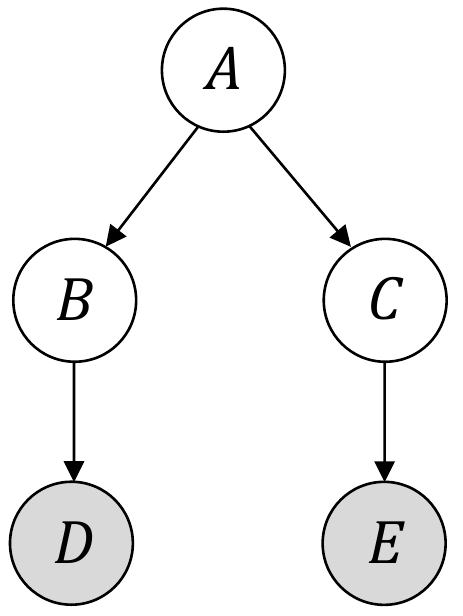}
        \caption{}
        \label{fig:example1-bn}
    \end{subfigure}
    \begin{subfigure}[b]{0.16\textwidth}
				\centering
        \includegraphics[scale=0.32]{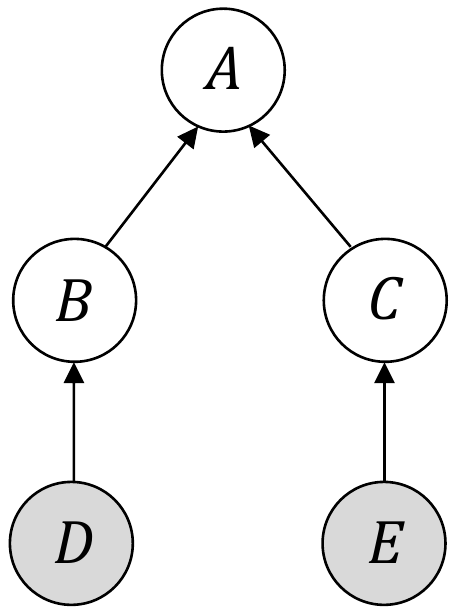}
        \caption{}
        \label{fig:example1-brooks}
    \end{subfigure}
    %\hfill
    \begin{subfigure}[b]{0.16\textwidth}
				\centering
        \includegraphics[scale=0.32]{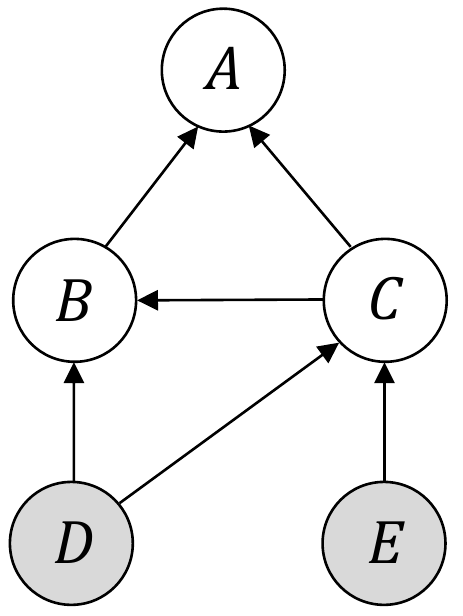}
        \caption{}
        \label{fig:example1-inverse}
    \end{subfigure}
		\begin{subfigure}[b]{0.16\textwidth}
				\centering
        \includegraphics[scale=0.32]{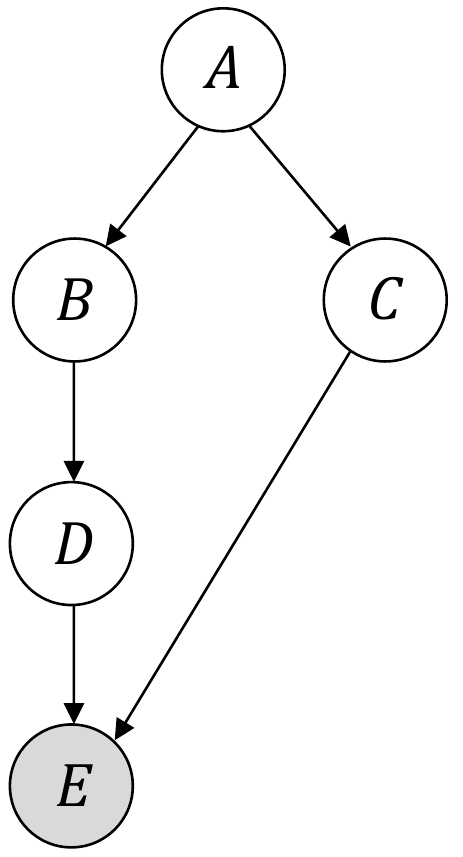}
        \caption{}
        \label{fig:example2-bn}
    \end{subfigure}
    \begin{subfigure}[b]{0.16\textwidth}
				\centering
        \includegraphics[scale=0.32]{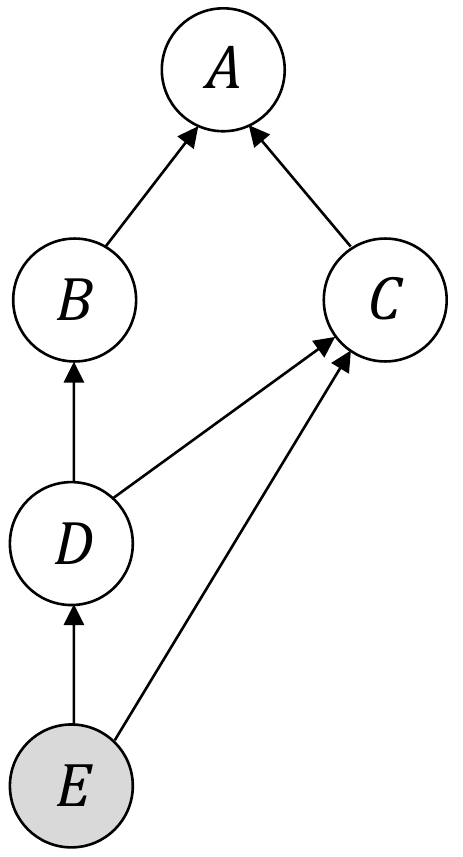}
        \caption{}
        \label{fig:example2-inverse}
    \end{subfigure}
		\begin{subfigure}[b]{0.16\textwidth}
				\centering
        \includegraphics[scale=0.32]{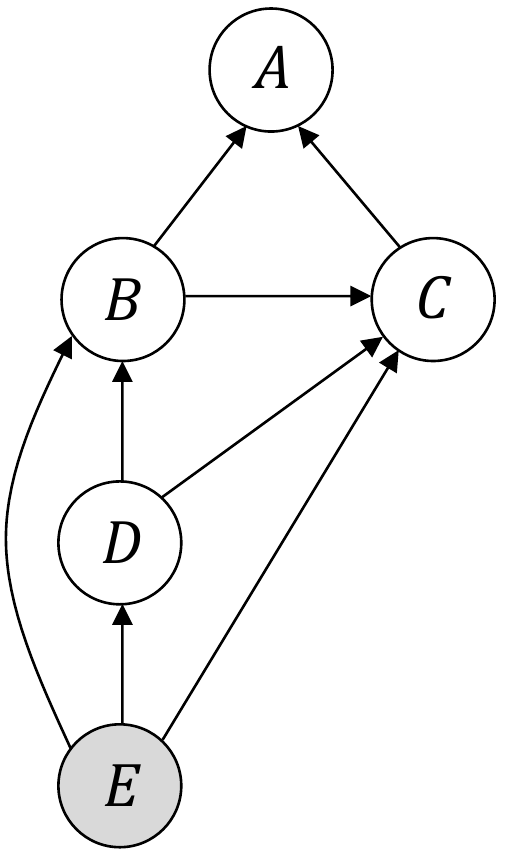}
        \caption{}
        \label{fig:example2-inverse}
    \end{subfigure}
    \caption[Counterexamples]{(a,d) Two simple BN structures for a generative model, (b,e) The corresponding inverse BN structures formed by Stuhlm{\"u}ller's Algorithm, (c,f) The inverse BN structure formed by our algorithm.
    This demonstrates how Stuhlm{\"u}ller's Algorithm can miss many edges and longer-term dependencies.}
    \label{fig:bn-examples}
\end{figure*}

\citet{StuhlmullerEtAl2013} give an algorithm for forming a ``heuristic inverse,'' $\mathcal{H}$, of a BN structure, $\mathcal{G}$.

First, let us define the concept of a Markov Blanket in a BN:
\begin{definition}
Let $\mathcal{G}$ be a BN structure over $\mathcal{X}$.
Then, the \emph{Markov blanket} of $X\in\mathcal{X}$ in $\mathcal{G}$, $\textnormal{Markov}_\mathcal{G}(X)$, is the minimal set of variables, $\mathbf{Z}$, that when conditioned on, make $X$ independent of $\mathcal{X}\setminus X$---that is, the set of parents, child, and parents of children of $X$.
\end{definition}
It is necessary to condition on the parents of a variable's children, because conditioning on its children may activate v-structures, and so we need to condition on the children's parents to block these paths.

Stuhlm\"{u}ller's algorithm works by visiting the variables of $\mathcal{G}$ in a reverse topological ordering, $Y_1,\ldots,Y_n$ (where $Y_i$ is equal to some observed $X_j$ or latent $Z_k$ depending on the structure of the graph and the ordering).
The graph $\mathcal{H}$ is produced by setting the parents of $Y_i$ to be the intersection of $Y_1,\ldots,Y_{i-1}$ and that node's Markov blanket in $\mathcal{G}$, excluding latent parents for observed nodes.
The procedure is equivalent to reversing the edges in $\mathcal{G}$, adding extra edges to fully connect all the parents of a node in $\mathcal{G}$, and removing edges from latent nodes into observed ones.
This produces the desired factorization $q(\mathbf{x}\mid\mathbf{z})q(\mathbf{z})$.

\citet{PaigeWood2016} claim that a heuristic inverse structure $\mathcal{H}$ is an I-map for $\mathcal{G}$, or equivalently, by the almost-everywhere completeness of d-separation, that $Y_1\rightleftharpoons\cdots\rightleftharpoons Y_m$ is active in $\mathcal{H}$ given $\mathbf{Z}$ implies that $Y_m\rightleftharpoons\cdots\rightleftharpoons Y_m$ is active in $\mathcal{G}$ given $\mathbf{Z}$, for an arbitrary trail.

If this were true, then we could factor $p$ as,
\begin{align*}
  p(\mathbf{y})
  &= \prod_{i=1}^np(y_i\mid y_1,\ldots,y_{i-1})\\
  &= \prod_{i=1}^np(y_i\mid \{y_1,\ldots,y_{i-1}\}
    \cap \textnormal{Markov}_\mathcal{G}(y_i)
    \cap \mathbb{I}(y_i)),
\end{align*}
where, $\mathbb{I}(y_i)=\mathbf{z}$ if $y_i\in\mathbf{z}$ and $\mathbf{y}$ otherwise, is defined to prevent edges from latent nodes into observed ones.

The problem is in going from the first to the second line.
For example, consider the factor for an arbitrary latent node, $Z_i$.
We have not conditioned on its \emph{complete} Markov blanket---only the children, and parents of children that occur previously in the ordering---and so we cannot assert that $Z_i$ is independent from all the other previous variables.

It is easy to construct counterexamples, for which the influence of a variable flows through one of its parents to effect another variable prior in the ordering that has not been conditioned on.
For instance, see Figure \ref{fig:bn-examples}.

Consider our first example in parts (a-b). The heuristic inverse, $\mathcal{H}$, in (b) asserts that $B\perp C$, since any path between the two variables is blocked by the v-structure. However, $B\perp C$ does not hold in the model, $\mathcal{G}$, in (a), as the path $B\leftarrow A\rightarrow C$ is active. As $\mathcal{H}$ asserts a conditional independence relationship that does not hold in $\mathcal{G}$, it is not faithful to the model. A similar argument can be produced for the second example in parts (d-e). A correct inverse structure produced by our algorithm is given in parts (c) and (f).

\section{Details of experimental setup}
Optimization was performed with Adam \citep{KingmaBa2014} and the default hyperparameters, $\beta_1=0.9$ and $\beta_2=0.999$.

\subsection{Relaxed Bernoulli VAEs}
We perform amortized SVI on a relaxed SBN with 30 latent units on the MNIST data set that has been statically binarized, and use the standard $50,000/10,000/10,000$ split for train/test/validation.
The relaxed Bernoulli prior had parameter $p=0.5$ and temperature $\tau=1/2$, and the relaxed Bernoulli distribution in the inference program, temperature $\tau=2/3$

A learning rate of \texttt{1e-4} was used, with batch size $100$.

In the forward model, $p(\mathbf{x}\mid\mathbf{z})$, the parameters were calculated by a tanh feedforward network with two hidden layers of size $[200,200]$.
For the ten mean-field inference programs, the same form of feedforward network was used, varying the size of the hidden layers from $[100,100]$, $[200,200]$,\ldots,$[1000,1000]$. The ten minimally faithful/fully connected inverses were parametrized similarly, adjusting upwards the size of the different hidden layers to match the number of parameters to the corresponding mean-field program.

The annealed importance sampling (AIS) estimate of $\ln(p(\mathbf{x}))$ averaged $5$ chains of $5000$ intermediate distributions.
As in \citet[C.3]{MaddisonEtAl2016}, the latents are treated in the logistic space rather than the relaxed Bernoulli space for numerical stability.
We found this was also essential for applying annealed importance sampling.

\subsection{Binary tree Gaussian BNs}\label{sec:binary-tree}
We model binary tree Gaussian BNs of depth $d$ with distribution, $X_0 \sim N(0,1)$, $X_i\mid x_{\left\lfloor(i-1)/2\right\rfloor}=y \sim N(w_iy, 1),\ \ \ i=1,\ldots,2^d-2$, where the $\{w_i\}$ are fixed constants sampled from $U[1/2,2]$ and we treat the leaves $\{x_{2^{d-1}-1},\ldots,x_{2^d-2}\}$ as the observed variables.

In both the heuristic/Stuhlm{\"u}ller's method and most compact inference programs, each inverse factor was parametrized with a normal distribution using a two hidden-layer ReLU feedforward network with $[100,100]$ and $[97,97]$ hidden units, respectively, to map from its parents to the distribution parameters.

A ReLU feedforward network with two hidden layers was also used for the fully connected and natural minimally faithful inference programs, with $[501,501]$ and $[1210,1210]$ hidden units, respectively. The MADE masks reduce the effective number of parameters, explaining why these numbers are greater than that for the heuristic inference program.

The total number of parameters for the heuristic, fully connected, most compact, and natural inference programs were $160545$, $159136$, $156021$, and $159901$, respectively.

The learning rate was initialized to {\ttfamily 1e-3}, decimating when learning converged, for example, every $100$ epochs in the case of $d=5$. A batchsize of $250$ was used, new samples from the generative model being drawn every minibatch for training, with 10 minibatches considered to constitute an epoch, and the test objective evaluated on a single minibatch every epoch.

The exact posterior under the true factorization can be calculated by using the equivalence between Gaussian BNs and multivariate normal distributions \citep[\S7.2]{KollerFriedman2009}---first the forward model is converted to the parameters of a multivariate normal distribution using Theorem 7.3, which is then transformed back into a Gaussian BN for the posterior using our true factorization and Theorem 7.4.
Samples from the posterior can be drawn by ancestral sampling.

We evaluate inference amortization by calculating the average log-posterior of a minibatch from the encoders every epoch under five fixed datasets of the observed variables (which have not be seen by the optimizer).

\subsection{Bayesian Gaussian Mixture Models}
We model a Bayesian Gaussian mixture model with $K=3$ clusters and $N=200$ two-dimensional samples. The variance parameters of the clusters were parametrized with $\sigma_{1k},\sigma_{2k},\rho_k$, where $\rho_k$ is the correlation between the two dimensions.  The inference network terms with distributions over vectors were parametrized by MADE, and each inverse factor was parametrized with a suitable probability distribution---$\phi$ with a Dirichlet, $\rho_k$ with Kumaraswamy, $\mu$ with a mixture of Gaussians, $\sigma_{1k}$ and $\sigma_{2k}$ with Inverse Gamma distributions, and $z$ with a Categorical.

The MADEs constituted of two hidden-layer ReLU feedforward network with 360 hidden units per layer for the NaMI inverse and 50 for the fully connected inverse, so that the total number of parameters in the network would be held fixed to allow for a fair comparison. The total number of parameters for the fully connected and natural inference programs were $820047$, and $826779$, respectively.

The learning rate was initialized to {\ttfamily 1e-3} and Adam algorithm was used. A dataset of $2000$ samples was sampled from the generative model for training the inference network, in minibatches of $200$. When the validation error decreased, a new dataset was drawn and training continued.

\subsection{Minimal and Non-minimal Faithful Inverses}
The setup for this experiment was as per \ref{sec:binary-tree} unless stated otherwise. We used a model of depth $d=4$ rather than $d=5$, parametrizing the forward-NaMI inverse with separate networks for each conditional distribution, rather than a single tree-MADE network. This was because adding extra edges to forward-NaMI broke the ability to share weights, and we wanted the same parametrization scheme for all three inverses. Each network had two hidden layers of size $100$. The two inverses with additional edges over the forward-NaMI one used networks with two hidden layers of size $99$ in order to keep the total capacity roughly fixed.

\section{Neural density estimators for weight-sharing}
\subsection{MADE}
We use the masked autoencoder distribution estimator (MADE) model \citep{GermainEtAl2015} extended for conditional distributions \citep{PaigeWood2016} to model fully connected distributions over latent variables, conditioning on all observations, that is,
\begin{align*}
	q(\mathbf{z}\mid\mathbf{x}) &= \prod^{m-1}_{i=0}q_i(z_i\mid z_1,\ldots,z_{i-1},\mathbf{x}).
\end{align*}
From a high level, MADE works by using a single feedforward network that takes as inputs $(\mathbf{x},\mathbf{z})$, and outputs parameters of all the factors $\{q_i\}$.
The weights of the feedforward network are multiplied elementwise by masking matrices so that if one were to trace a path back from an output parameter for $q_i$ to the inputs, that parameter would only be connected to $\{z_1,\ldots,z_{i-1},\mathbf{x}\}$.

To make things more concrete, consider a single-hidden-layer feedforward network, used to the calculated the parameters, $\theta$, of binary valued data,
\begin{align*}
	\mathbf{h} &= \sigma_w\left(\mathbf{b} + (W\odot M^{(w)})(\mathbf{z},\mathbf{x})\right)\\
	\theta &= \sigma_v\left(\mathbf{c} + (V\odot M^{(v)})\mathbf{h}\right),
\end{align*}
where $\mathbf{b},\mathbf{c},W,V$ are real-valued parameters to be learned, $\odot$ denotes elementwise multiplication, $\sigma_w,\sigma_v$ are nonlinear functions, and $M_w,M_v$ are fixed binary masks.

To each hidden unit, $h_i$, we assign an integer uniformly from $\{1,\ldots,m-1\}$.
To each input unit we assign the integer $0$ if it corresponds to an observation, $x_i$, and the integer $i$ if it corresponds to the latent unit $z_j$.
The input mask element $M^{(w)}_{i,j}$ represent a connection from the $i$th input unit to the $j$th hidden unit.
Thus we set $M^{(w)}_{i,j}=1$ only when the integer assigned to the $i$th input is less than the integer assigned to the $j$th hidden unit, and $0$ otherwise. In this way, if the $j$th hidden unit is assigned $k$, it will depend on $\{z_1,\ldots,z_{k-1}, \mathbf{x}\}$.
The output mask $M^{(v)}$ is constructed similarly by assigning the integer $i$ the units corresponding to the parameters of $q_i$.

This method can be easily extended to feedforward networks with more than one hidden layer.
For instance, if there is a second hidden layer $\mathbf{h}'$ with mask $M^{(w')}$, we assign each hidden unit $h'_i$ an integer uniformly from $\{1,\ldots,m-1\}$ (or in fact, we can start from the lowest integer assigned to an $h_i$), and set $M^{(W')}_{i,j}=1$ only when the integer assigned to $h_i$ is less than or equal to the integer assigned to $h'_j$.
In this way, if $h'_j$ is assigned integer $k$, it depends on $\{z_1,\ldots,z_{k-1},\mathbf{x}\}$ through hidden units $\{h_i\}$ assigned $k$, it depends on $\{z_1,\ldots,z_k-2\}$ through hidden units $\{h_i\}$ assigned $k-1$, and so on.
This is a form of weight sharing.

We use two hidden layer MADEs in our experiments, including, in addition, masked skip-weights from the inputs to the outputs, as is recommended in \citet{GermainEtAl2015}.

\subsection{Tree MADE}
In trying to model the regular but less-than-fully-connected dependency structure of minimally faithful inverses to binary trees, we had the following novel insight.
Rather than thinking of the integers assigned to the input, hidden, and output units as simply numbers, we recognize that they actually identify subsets of the model variables.
That is, $k$ corresponds to $\{z_0,\ldots,z_{k-1},\mathbf{x}\}$.
The mask weight is set to $1$ only when the first subset is contained in the second.
A difference choice of subsets will allow us to model another dependency structure, with the subset inclusion relationship defining weight sharing across the factors.

Running our algorithm on the binary tree Gaussian network of $\S3.2$, reveals that one minimally faithful inverse for a model of depth $d$ comprises factors,
\begin{align*}
	q_i(x_i\mid x_{i+1},\ldots,x_{2(i+1)}),\ \ \ i=0,1,\ldots,2^d-2.
\end{align*}
We break up the subsets $\{x_{i+1},\ldots,x_{2(i+1)}\}$ into,
\begin{align*}
	& \{x_{i+1}\},\\
	& \{x_{i+2},\ldots,x_{2(i+1)}\},\\
	& \{x_{i+3},\ldots,x_{2(i+1)}\},\\
	& \vdots\\
	& \{x_{2i+1},\ldots,x_{2(i+1)}\}
\end{align*}
	for each $i$, and assign each a unique integer.
The hidden units are uniformly assigned one of these subsets.
The input unit for $x_i$ is assigned the subset $\{x_i\}$ and the output units for the parameters of $q_i$ are assigned the subset $\{x_{i+1}\ldots,x_{2(i+1)}\}$.
The mask from one hidden, input, or output unit to another is set to $1$ only when the subset corresponding to the first unit is contained in the subset corresponding to the second unit.

By construction, this feedforward network will give the parameters for the $\{q_i\}$ such that $q(\mathbf{z}\mid\mathbf{x})$ is a minimal I-map for the posterior.
This idea can clearly be generalized to arbitrary dependency structures, which we leave for future work.
We can algorithmically determine the form of the inverse factors in a minimally faithful inverse offline, extract all subsets of their scopes, and perform the same procedure as above.

\input{background}

\vfill\pagebreak
\bibliography{longstrings,dphil}
\bibliographystyle{icml2018}

%% file: preamble.tex
\newtheorem{theorem}{Theorem}

\newtheorem{definition}{Definition}

%% file: abstract.tex
Inference amortization methods share information across multiple posterior-inference problems, allowing each to be carried out more efficiently.
Generally, they require the inversion of the dependency structure in the generative model, as the modeller must learn a mapping from observations to distributions approximating the posterior.
Previous approaches have involved inverting the dependency structure in a heuristic way that fails to capture these dependencies correctly, thereby limiting the achievable accuracy of the resulting approximations.
We introduce an algorithm for faithfully, and minimally, inverting the graphical model structure of any generative model.
Such inverses have two crucial properties:
\begin{inparaenum}[a)]
\item they do not encode any independence assertions that are absent from the model and
\item they are local maxima for the number of true independencies encoded.
\end{inparaenum}
We prove the correctness of our approach and empirically show that the resulting minimally faithful inverses lead to better inference amortization than existing heuristic approaches.

%%% Local Variables:
%%% mode: latex
%%% TeX-master: "main"
%%% End:

%% file: introduction.tex
\section{Introduction}%
\label{sec:intro}
Evidence from human cognition suggests that the brain reuses the results of past inferences to speed up subsequent related queries \citep{Gershman2014}.
In the context of Bayesian statistics, it is reasonable to expect that, given a generative model, $p(\mathbf{x},\mathbf{z})$, over data $\mathbf{x}$ and latent variables $\mathbf{z}$, inference on $p(\mathbf{z}\mid\mathbf{x}_1)$ is informative about inference on $p(\mathbf{z}\mid\mathbf{x}_2)$ for two related inputs, $\mathbf{x}_1$ and $\mathbf{x}_2$.
Several algorithms \citep{KingmaWelling2013, RezendeEtAl2014, StuhlmullerEtAl2013, PaigeWood2016, LeEtAl2016, LeEtAl2017, MaddisonEtAl2017, NaessethEtAl2017} have been developed with this insight to perform \emph{amortized inference} by learning an inference artefact $q(\mathbf{z}\mid\mathbf{x})$, which takes as input the values of the observed variables, and---typically with the use of neural network architectures---return a distribution over the latent variables approximating the posterior.
These inference artefacts are known variously as inference networks, recognition models, probabilistic encoders, and guide programs; we will adopt the term \emph{inference networks} throughout.

Along with conventional fixed-model settings~\citep{StuhlmullerEtAl2013,LeEtAl2016,Ritchie2016,PaigeWood2016},
a common application of inference amortization is in the training of variational auto-encoders (VAEs)~\citep{KingmaWelling2013}, for which the inference network is simultaneously learned alongside a generative model.
It is well documented that deficiencies in the expressiveness or training of the inference network can also have a knock-on effect on the learned generative model in such contexts~\citep{burda2015importance,cremer2017reinterpreting,cremer2018inference,rainforth2018tighter}, meaning that poorly chosen coarse-grained structures can be particularly damaging.

Implicit in the factorization of the generative model and inference network in both fixed and learned model settings are probabilistic graphical models, commonly Bayesian networks (BNs), encoding dependency structures. We refer to these as the \emph{coarse-grain} structure, in opposition to the \emph{fine-grain} structure of the neural networks that form each inference (and generative) network factor.
In this sense, amortized inference can be framed as the problem of graphical model inversion---how to invert the graphical model of the generative model to give a graphical model approximating the posterior. Many models from the deep generative modeling literature can be represented as BNs \citep{KrishnanEtAl2017, GanEtAl2015, Neal1990, KingmaWelling2013, GermainEtAl2015, VanDenOordEtAl2016a, VanDenOordEtAl2016b}, and fall within this framework.

In this paper, we borrow ideas from the probabilistic graphical models literature, to address the previously open problem of how best to automate the design of the coarse-grain structure of the inference network \citep{Ritchie2016}.
Typically, the inverse graphical model is formed heuristically.
At the simplest level, some methods just invert the edges in the BN for the generative model, removing edges between observed variables \citep{KingmaWelling2013, GanEtAl2015, RanganathEtAl2014}.
In a more principled, but still heuristic, approach, \citet{StuhlmullerEtAl2013,PaigeWood2016} construct the inference network by inverting the edges and additionally connecting the parents of children in the original graph (both of which are a subset of a variable's Markov blanket; see Appendix C).\input{fig-student-graphs} \vspace{-10pt}

In general, these heuristic methods introduce conditional independencies into the inference network that are not present in the original distribution.  Consequently, they cannot represent the true posterior
even in the limit of infinite neural network capacities.
Take the simple generative model with branching structure of Figure \ref{fig:example1-bn}.
The inference network formed by Stuhlm\"{u}ller's method inverts the edges of the model as in Figure \ref{fig:example1-brooks}.
However, an inference network that is able to represent the true posterior requires extra edges between the branches, as in Figure \ref{fig:example1-inverse}.

Another approach, taken by \citet{LeEtAl2016}, is to use a fully connected BN for the inverse graphical model, such that
every random choice made by the inference network depends on every previous one.
Though such a model is expressive enough to correctly represent the data given infinite capacity and training time, it ignores substantial available information from the forward model, inevitably leading to reduced performance for finite
training budgets and/or network capacities.

In this paper, we develop a tractable framework to remedy these deficiencies:
 the
\textbf{\emph{Na}}tural \textbf{\emph{M}}inimal \textbf{\emph{I}}-map generator (NaMI).
Given an arbitrary BN structure, NaMI can be used to construct an inverse BN structure that is provably both \emph{faithful} and \emph{minimal}.
It is faithful in that it contains sufficient edges to avoid encoding conditional independencies absent from the model.
It is minimal in that it does not contain any unnecessary edges; i.e., removing any edge would result in an unfaithful structure.

NaMI chiefly draws upon variable elimination \citep[Ch 9,10]{KollerFriedman2009}, a well-known algorithm from the graphical model literature for performing exact inference on discrete factor graphs.
The key idea in the operation of NaMI is to simulate variable elimination steps as a tool for successively determining a minimal, faithful, and natural inverse structure, which can then be used to parametrize an inference network.
NaMI further draws on ideas such as the min-fill heuristic \citep{FishelsonGeiger2004}, to choose the ordering in which variable elimination is simulated, which in turn influences the structure of the generated inverse.

To summarize, our key contributions are:
\begin{compactenum}[i)]
\item framing generative model learning through amortized variational inference as a graphical model inversion problem, and
\item using the simulation of exact inference algorithms to construct an algorithm for generating provably minimally faithful inverses.
\end{compactenum}
Our work thus highlights the importance of constructing both minimal and faithful inverses, while providing the first approach to produce inverses satisfying these properties.

%%% Local Variables:
%%% mode: latex
%%% TeX-master: "main"
%%% End:

%% file: fig-student-graphs.tex
\begin{wrapfigure}[12]{r}{0.35\textwidth}
 \vspace{-5pt}
  \centering
  \subcaptionbox{\label{fig:example1-bn}}%
  {\includegraphics[width=0.29\linewidth]{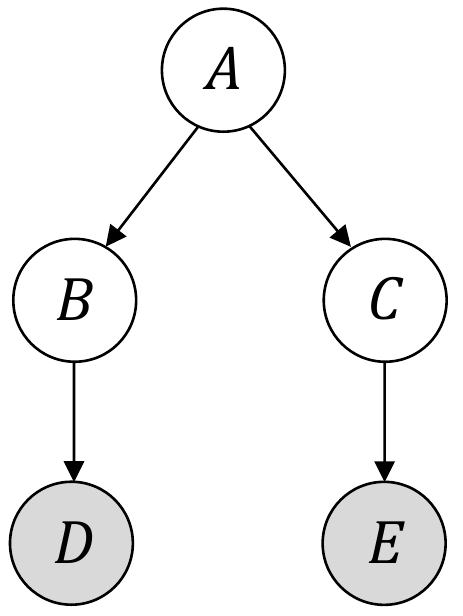}}
  \,
  \subcaptionbox{\label{fig:example1-brooks}}%
  {\includegraphics[width=0.29\linewidth]{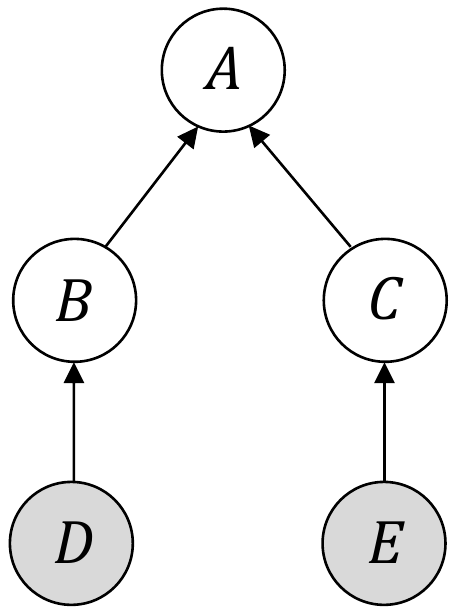}}
  \,
  \subcaptionbox{\label{fig:example1-inverse}}%
  {\includegraphics[width=0.29\linewidth]{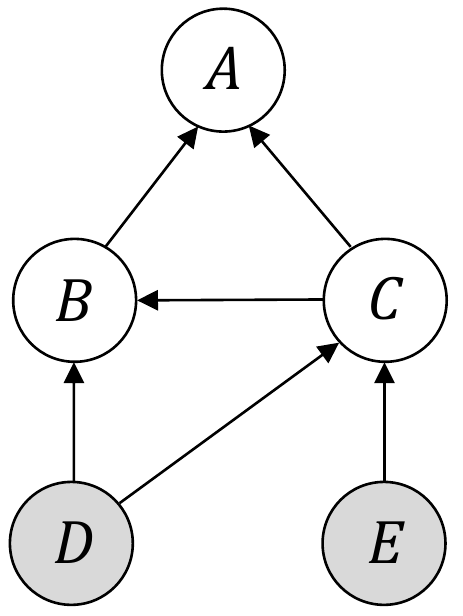}}%
  \vspace*{-1.2ex}
	\caption[Simple examples]{(a) Generative model BN; (b) Inverse BN by Stuhlm{\"u}ller's Algorithm; (c) \emph{Faithful} inverse BN by our algorithm.
	\label{fig:student-graphs}}
\end{wrapfigure}

%% file: method.tex
\section{Method}
Our algorithm builds upon the tools of \emph{probabilistic graphical models}--- a summary for unfamiliar readers is given in Appendix A.

\subsection{General idea}\label{sec:stochastic-inverses}
Amortized inference algorithms make use of inference networks that approximate the posterior.
To be able to represent the posterior accurately, the distribution of the inference network should not encode independence assertions that are absent from the generative model.
An inference network that did encode additional independencies could not represent the true posterior, even in the non-parametric limit, with neural network factors whose capacity approaches infinity.

Let us define a \emph{stochastic inverse} for a generative model $p(\mathbf{x}|\mathbf{z})p(\mathbf{z})$ that factors according to a BN structure $\mathcal{G}$ to be a factorization of $q(\mathbf{z}|\mathbf{x})q(\mathbf{x})$ over $\mathcal{H}$ \citep{StuhlmullerEtAl2013,PaigeWood2016}.
The $q(\mathbf{z}|\mathbf{x})$ part of the stochastic inverse will define the factorization, or rather, coarse-grain structure, of the inference network.
Recall from \S\ref{sec:intro} that this involved two characteristics.
We first require $\mathcal{H}$ to be an \emph{I-map} for $\mathcal{G}$:
\begin{definition}
	Let $\mathcal{G}$ and $\mathcal{H}$ be two BN structures.
Denote the set of all conditional independence assertions made by a graph, $\mathcal{K}$, as $\mathcal{I}(\mathcal{K})$.
We say $\mathcal{H}$ is an \emph{I-map} for $\mathcal{G}$ if $\mathcal{I}(\mathcal{H})\subseteq\mathcal{I}(\mathcal{G})$.
\end{definition}
To be an I-map for $\mathcal{G}$, $\mathcal{H}$ may not encode all the independencies that $\mathcal{G}$ does, but it must not mislead us by encoding independencies not present in $\mathcal{G}$.
We term such inverses as being \emph{faithful}.
While the aforementioned heuristic methods \emph{do not} in general produce faithful inverses, using either a fully-connected inverse, or our method, does.

Second, since a fully-connected graph encodes no conditional independencies and is therefore suboptimal, we require in addition that $\mathcal{H}$ be a \emph{minimal I-map} for $\mathcal{G}$:
\begin{definition}
	A graph $\mathcal{K}$ is a \emph{minimal I-map} for a set of independencies $\mathcal{I}$ if it is an I-map for $\mathcal{I}$ and if removal of even a single edge from $\mathcal{K}$ renders it not an I-map.
\end{definition}
We call such inverses \emph{minimally faithful}, which roughly means that the inverse is a local optimum in the number of true independence assertions it encodes.

There will be many minimally faithful inverses for $\mathcal{G}$, each with a varying number of edges.
Our algorithm produces a \emph{natural inverse} in the sense that it either inverts the order of the random choices from that of the generative model (when it is run in the topological mode), or it preserves the ordering \input{fig-natural} of the random choices (when it is run in reverse topological mode):
\begin{definition}\label{def:naturalness}
	A stochastic inverse $\mathcal{H}$ for $\mathcal{G}$ over variables $\mathcal{X}$ is  a \emph{natural inverse} if either, for all $X\in\mathcal{X}$ there are no edges in $\mathcal{H}$ from $X$ to its descendants in $\mathcal{G}$, or, for all $X\in\mathcal{X}$ there are no edges in $\mathcal{H}$ from $X$ to its ancestors in $\mathcal{G}$.
\end{definition}
Essentially, a natural inverse is one for which if we were to perform ancestral sampling, the variables would be sampled in either a topological or reverse-topological ordering, relative to the original model. Consider the inverse networks
of $\mathcal{G}$ shown in Figure \ref{fig:natural-graphs}. $\mathcal{H}_1$ is not a natural inverse of $\mathcal{G}$, since there is both an edge $A\rightarrow C$ from a parent to a child, and an edge $C\rightarrow B$ from a child to a parent, relative to $\mathcal{G}$. However, $\mathcal{H}_2$ and $\mathcal{H}_3$ are natural, as they correspond respectively to the reverse-topological and topological orderings $C,B,A$ and $B,A,C$.

Most heuristic methods, including those of~\citep{StuhlmullerEtAl2013,PaigeWood2016}, produce (unfaithful) natural inverses that invert the order of the random choices, giving a reverse-topological ordering.

\subsection{Obtaining a natural minimally faithful inverse}
\label{sec:true-factorization}

\begin{wrapfigure}[8]{r}{0.26\textwidth}
  \vspace*{-2ex}
	\centering
	\includegraphics[scale=0.34]{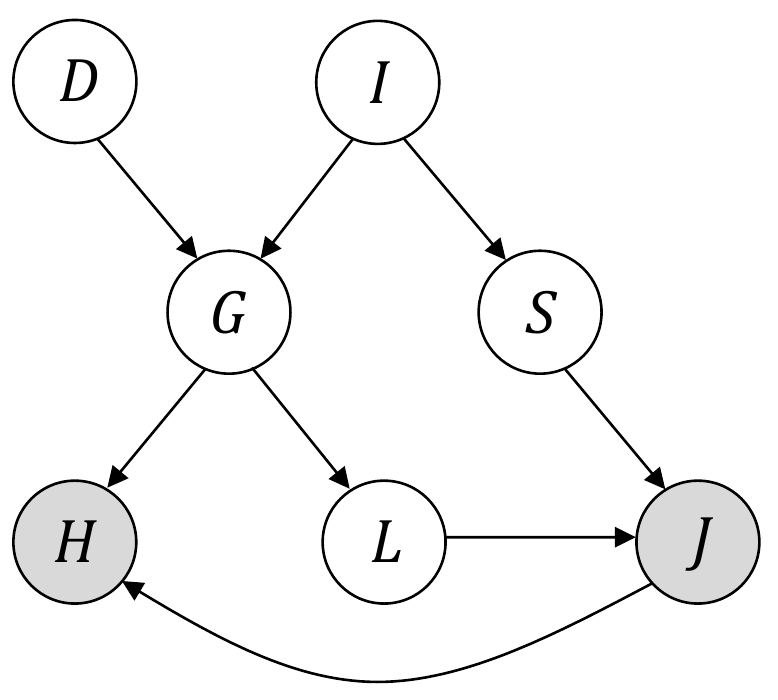}
	\caption{Example BN}
  \label{fig:student-bn}
\end{wrapfigure}

We now present NaMI's graph inversion procedure that given an arbitrary BN structure, $\mathcal{G}$, produces a natural minimal I-map, $\mathcal{H}$.
We illustrate the procedure step-by-step on the example given in Figure \ref{fig:student-bn}. Here $H$ and $J$ are observed, as indicated by the shaded nodes.
Thus, our latent variables are $\mathbf{Z}=\{D,I,G,S,L\}$, our data is $\mathbf{X}=\{H,J\}$, and a factorization for $p(\mathbf{z}\mid\mathbf{x})$ is desired.

The NaMI graph-inversion algorithm is traced in Table \ref{tab:inversion-example}.
Each step incrementally constructs two graphs:
an \emph{induced graph} $\mathcal{J}$ and a \emph{stochastic inverse} $\mathcal{H}$.
The induced graph is an undirected graph whose maximally connected subgraphs, or \emph{cliques}, correspond to the scopes of the intermediate factors produced by simulating variable elimination.
The stochastic inverse represents our eventual target which encodes the inverse dependency structure.  It is constructed using information from the partially-constructed induced graph.
Specifically, NaMI goes through the following steps for this example.

\textbf{{\scshape Step 0}}: The partial induced graph and stochastic inverse are initialized.
The initial induced graph is formed by taking the directed graph for the forward model, $\mathcal{G}$, removing the directionality of the edges, and adding additional edges between variables that share a child in $\mathcal{G}$---in this example, edges $D-I$, $S-L$ and $G-J$.
This process is known as \emph{moralization}.
The stochastic inverse begins as disconnected variables, and edges are added to it at each step.
\input{tab-inversion-example}

\textbf{{\scshape Step 1}}: The frontier set of variables to consider for elimination, $S$, is initialized to the latent variables having no latent parents in $\mathcal{G}$, that is, $D$, $I$.
To choose which variable to eliminate first, we apply the greedy min-fill heuristic, which is to choose the (possibly non-unique) variable that adds the fewest edges to the induced graph $\mathcal{J}$ in order to produce as compact an inverse as possible under the topological ordering.
Specifically, noting that the cliques of $\mathcal{J}$ correspond to the scopes of intermediate factors during variable elimination, we want to avoid producing intermediate factors which would require us to add additional edges to $\mathcal{J}$, as doing so will in turn induce additional edges in $\mathcal{H}$ at future steps.
For this example, if we were to eliminate $D$, that would produce an intermediate factor, $\psi_D(D,I,G)$, while if we were to eliminate $I$, that would produce an intermediate factor, $\psi_I(I,D,G,S)$.
Choosing to eliminate would $I$ thus requires adding an edge $G\hbox{--}S$ to the induced graph, as there is no clique $I,D,G,S$ in the current state of $\mathcal{J}$.
Conversely, eliminating $D$ does not require adding extra edges to $\mathcal{J}$ and so we choose to eliminate $D$.

 The elimination of $D$ is simulated by marking its node in $\mathcal{J}$.
The parents of $D$ in the inverse $\mathcal{H}$ are set to be its  nonmarked neighbours in $\mathcal{J}$, that is, $I$ and $G$.
$D$ is then removed from the frontier,  and any non-observed children in $\mathcal{G}$ of $D$ whose parents have all been marked added to it---in this case, there are none as the only child of $D$, $G$, still has an unmarked parent $I$.

\textbf{{\scshape Step 2}}: Variable $I$ is the sole member of the frontier and is chosen for elimination.
The elimination of $I$ is simulated by marking its node in $\mathcal{J}$ \emph{and} adding the additional edge $G\hbox{--}S$.
This is required because elimination of $I$ requires the addition of a factor, $\psi_I(I,G,S)$, that is not currently present in $\mathcal{J}$.
The parents of $I$ in the inverse $\mathcal{H}$ are set to be its nonmarked neighbours in $\mathcal{J}$, $G$ and $S$.
$I$ is then removed from the frontier.
Now, $G$ and $S$ are children of $I$, and both their parents $D$ and $I$	have been marked.
Therefore, they are added to the frontier.

\textbf{{\scshape Step 3-5}}: The process is continued until the end of the fifth step when all the latent variables, $D,I,S,G,L$, have been eliminated and the frontier is empty.
At this point, $\mathcal{H}$ represents a factorization $p(\mathbf{z}\mid\mathbf{x})$, and we stop here as only a factorization for the posterior is required for amortized inference. Note, however, that it is possible to continue simulating steps of variable elimination on the observed variables to complete the factorization as $p(\mathbf{z}\mid\mathbf{x})p(\mathbf{x})$.

An important point to note is that NaMI's graph inversion can be run in one of two modes. The ``topological mode,'' which we previously implicitly considered, simulates variable elimination in a topological ordering, producing an inverse that reverses the order of the random choices from the generative model.
Conversely, NaMI's graph inversion can also be run in ``reverse topological
\input{alg-invert-bn}
mode,'' which simulates variable elimination in a reverse topological ordering, producing an inverse that preserves the order of random choices in
the generative model.
We will refer to these approaches as \emph{forward-NaMI} and \emph{reverse-NaMI} respectively in the rest of the paper.
The rationale for these two modes is that, though they both produce minimally faithful inverses, one may be substantially more compact than the other, remembering that minimality only ensures a local optimum.
For an arbitrary graph, it cannot be said in advance which ordering will produce the more compact inverse.
However, as the cost of running the inversion algorithm is low, it is generally feasible to try and pick the one producing a better solution.

 The general NaMI graph-reversal procedure is given in Algorithm \ref{alg:invert-bn}.
It is further backed up by the following formal demonstration of correctness, the proof for which is given in Appendix F.
\vspace{-2pt}\begin{theorem}\label{theorem:correctness}
	The Natural Minimal I-Map Generator of Algorithm 1 produces inverse factorizations that are natural and minimally faithful.
\end{theorem}\vspace{-4pt}

We further note that NaMI's graph reversal has a running time of order $O(nc)$ where $n$ is the number of 
latent variables in the graph and $c<<n$ is the size of the largest clique in the induced graph.  We consequently see that it can be run
cheaply for practical problems: the computational cost of generating the
inverse is generally dominated by that of training the resulting
inference network itself.  See Appendix F for more details.

\subsection{Using the faithful inverse}
Once we have obtained the faithful inverse structure $\mathcal{H}$, the next step is to use it to learn an inference network, $q_\psi(\mathbf{z}\mid\mathbf{x})$. For this, we use the factorization given by $\mathcal{H}$.
Let $\tau$ denote the reverse of the order in which variables were selected for elimination by Line 9 in Algorithm 1, 
such that $\tau$ is a permutation of $1,\dots,n$ and $\tau(n)$ is the first variable eliminated.  
$\mathcal{H}$ encodes the factorization
\begin{align}
q_\psi(\mathbf{z}\mid\mathbf{x})=\prod\nolimits^n_{i=1} q_{i}(z_{\tau(i)}\mid\text{Pa}_\mathcal{H}(z_{\tau(i)}))
\end{align}
where $\text{Pa}_\mathcal{H}(z_{\tau(i)})\subseteq\left\{\mathbf{x},z_{\tau(1)},\dots,z_{\tau(i-1)}\right\}$ indicates the parents of $z_{\tau(i)}$ in $\mathcal{H}$.
For each factor $q_{i}$, we must decide both the class of distributions for $z_{\tau(i)}\mid\text{Pa}_\mathcal{H}(z_{\tau(i)})$, and how the parameters for that class are calculated.
Once learned, we can both sample from, and evaluate the density of, the inference network for a given dataset by considering each factor in turn.

The most natural choice for the class of distributions for each factor is to use
the same distribution family as the corresponding variable in the generative model, such that the supports of these distributions match.
For instance, continuing the example from Figure~\ref{fig:student-bn}, if $D\sim N(0,1)$ in the generative model, then
a normal distribution would also be used for
$D\mid I,G$ in the inference network. To establish the mapping from data to the parameters to this distribution,
we train neural networks using stochastic gradient ascent methods.  For instance, we could set
$D\mid\{I=i,G=g\}\sim N(\mu_\varphi(i,g),\sigma_\varphi(i,g))$, where $\mu_\varphi$ and $\sigma_\varphi$ are two densely connected
feedforward networks, with learnable parameters $\varphi$.
In general, it will be important to choose architectures which well match the problem at hand.
For example, when perceptual inputs such as images and language are present in the conditioning variables, it is
advantageous to first embed them to a lower-dimensional representation using, for example, convolutional neural networks.

Matching the distribution families in the inference network and generative model, whilst a simple and often adequate approximation, can be suboptimal. For example, suppose that for a normally distributed variable in the generative model, the true conditional distribution in the posterior for that variable is multimodal. In this case, using a (single mode) normal factor in the inference network would not suffice. One could straightforwardly instead use, for example, either a mixture of Gaussians, or, normalizing flows \citep{RezendeMohamed2015,KingmaEtAl2016}, to parametrize each inference network factor in order to improve expressivity, at the cost of additional implementational complexity.
In particular, if one were to use a provably universal density estimator to parameterize each
inference network factor, such as that introduced in~\citet{HuangEtAl2018}, the resulting NaMI inverse
would constitute a universal density estimator of the true posterior.

After the inference network has been parametrized, it can be trained in number of different ways, depending on the final use
case of the network.  For example, in the context of amortized stochastic variational inference (SVI) methods such as VAEs
\citep{KingmaWelling2013,RezendeEtAl2014}, the model $p_\theta(\mathbf{x},\mathbf{z})$ is learned along with the inference network $q_\psi(\mathbf{z}\mid\mathbf{x})$ by optimizing a lower bound on the marginal loglikelihood of the data, $\mathcal{L}_{\sc ELBO} = \mathbb{E}_{q_\psi(\mathbf{z}|\mathbf{x})}\left[\ln p_\theta(\mathbf{x},\mathbf{z}) - \ln q_\psi(\mathbf{z}\mid\mathbf{x})\right]$.
Stochastic gradient ascent can then be used to optimize $\mathcal{L}_{\sc ELBO}$ in the same way a standard VAE,
simulating from $q_\psi (z|x)$ by considering each factor in turn and using reparameterization~\citep{KingmaWelling2013}
when the individual factors permit doing so.

A distinct training approach is provided when the model $p(\mathbf{x},\mathbf{z})$ is fixed \citep{PapamakariosMurray2015}.
Here a proposal is learnt for either importance sampling~\citep{LeEtAl2016}
or sequential Monte Carlo~\citep{PaigeWood2016} by using stochastic gradient ascent to
minimize the reverse KL-divergence between the inference network $q_\psi(\mathbf{z}\mid\mathbf{x})$ and the true posterior
$p(\mathbf{z}\mid\mathbf{x})$. Up to a constant, the objective is given by
$\mathcal{L}_{\sc IC} = \mathbb{E}_{p(\mathbf{x},\mathbf{z})}\left[-\ln q_\psi(\mathbf{z}\mid\mathbf{x})\right].$

Using a minimally faithful inverse structure typically improves the best inference network attainable and the finite time training performance for both these settings, compared with previous naive approaches.
In the VAE setting, this can further have a knock-on effect on the quality of the learned model $p_\theta(\mathbf{x},\mathbf{z})$, both because a better inference network will give lower variance updates of the generative network~\citep{rainforth2018tighter} and because restrictions in the expressiveness of the inference network lead to similar restrictions in the generative network~\citep{cremer2017reinterpreting,cremer2018inference}.

In deep generative models, the BNs may be much larger than the examples shown here. However, typically at the macro-level, where we collapse each vector to a single node, they are quite simple. When we invert this type of collapsed graph, we must do so with the understanding that the distribution over a vector-valued node in the inverse must express dependencies between all its elements in order for the inference network to be faithful.

%%% Local Variables:
%%% mode: latex
%%% TeX-master: "main"
%%% End:

%% file: fig-natural.tex
\begin{wrapfigure}[15]{r}{0.25\textwidth}
  \centering\vspace*{-4pt}
  \subcaptionbox*{\label{fig:natural-bn}$\mathcal{G}$}%
  {\includegraphics[width=0.44\linewidth]{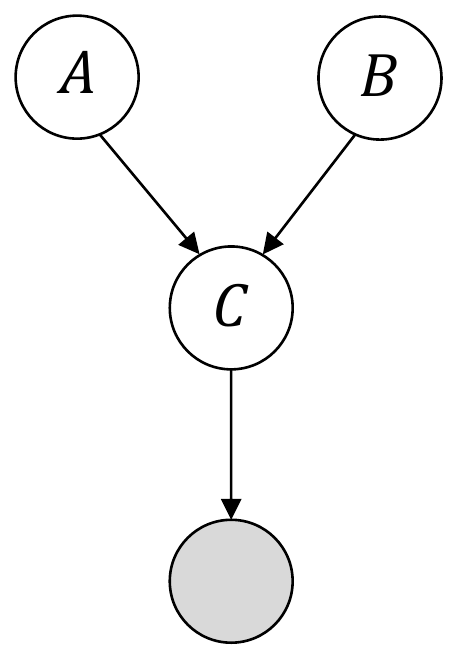}}
  \,
  \subcaptionbox*{\label{fig:natural-unnatural}$\mathcal{H}_1$}%
  {\includegraphics[width=0.44\linewidth]{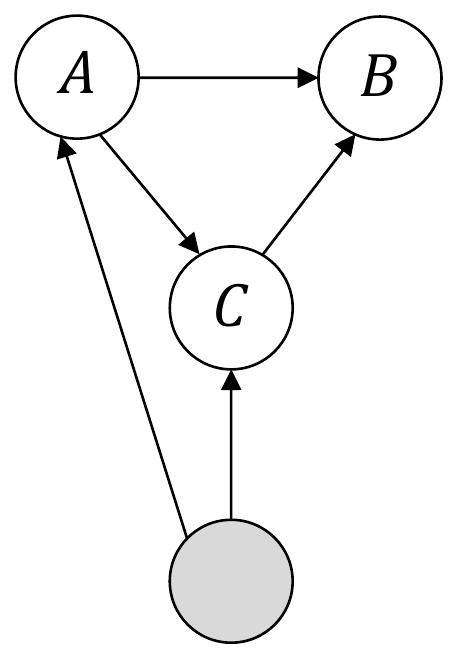}}
  \vspace{4pt}
  \subcaptionbox*{\label{fig:natural-forward}$\mathcal{H}_2$}%
  {\includegraphics[width=0.44\linewidth]{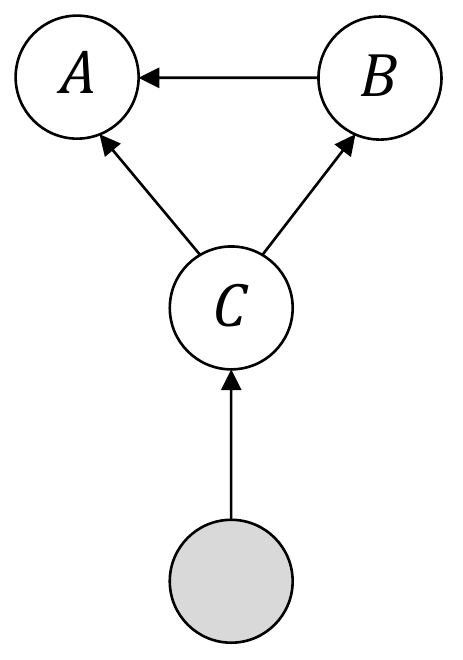}}
	\,
  \subcaptionbox*{\label{fig:natural-reverse}$\mathcal{H}_3$}%
  {\includegraphics[width=0.44\linewidth]{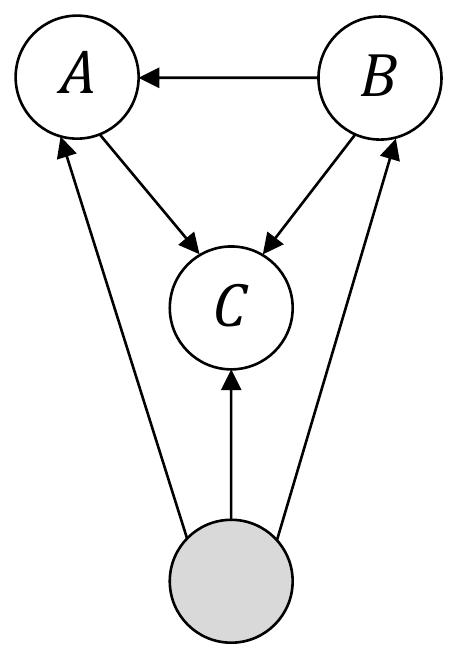}}
  \vspace*{-1.2ex}
	\caption[Illustrating naturalness]{Illustrating definition of naturalness.
	\label{fig:natural-graphs}}
\end{wrapfigure}

%% file: tab-inversion-example.tex
\begin{table}[t]
	\caption{Tracing the NaMI algorithm on example from Figure~\ref{fig:student-bn}. $S$ is the set of ``frontier'' variables that are considered for elimination, $v\in S$ the variable eliminated at each step chosen by the greedy min-fill heuristic, $\mathcal{J}$ the partially constructed induced graph \emph{after} each step with black nodes indicating a eliminated variables, and $\mathcal{H}$ the partially constructed stochastic inverse.}
\begin{minipage}{0.40\textwidth}
		\centering
		\vspace{5pt}
	\begin{tabular}{
		@{\hspace*{0ex}}m{0.14\linewidth}
		@{\hspace*{1ex}}m{0.05\linewidth}
		@{\hspace*{5ex}}m{0.05\linewidth}
		@{\hspace*{3ex}}m{0.30\linewidth}
		@{\hspace*{3ex}}m{0.30\linewidth}}
	\toprule
{\small\scshape Step}~ & ~$S$ & $v$ & ~~~~~~~$\mathcal{J}$ & ~~~~~~~~$\mathcal{H}$ \\
	\midrule
	~0 & $\emptyset$ & $\emptyset$ & \includegraphics[scale=0.24]{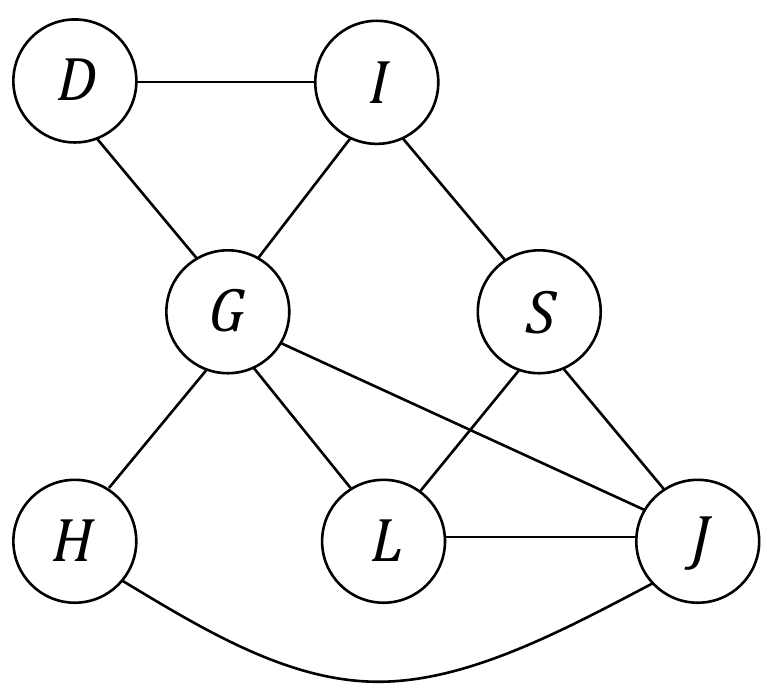} & \includegraphics[scale=0.24]{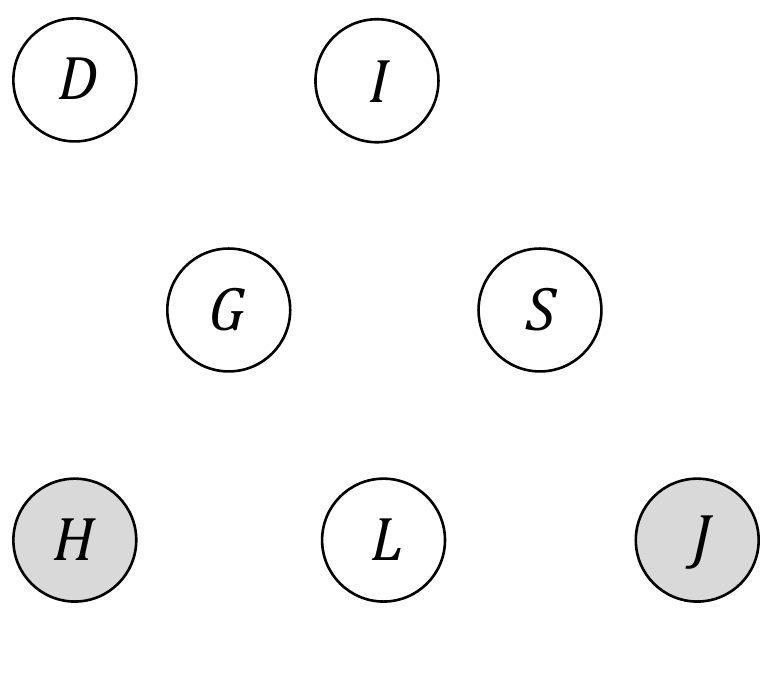} \\
	\midrule
	~1 & $D,I$ & $D$ & \includegraphics[scale=0.24]{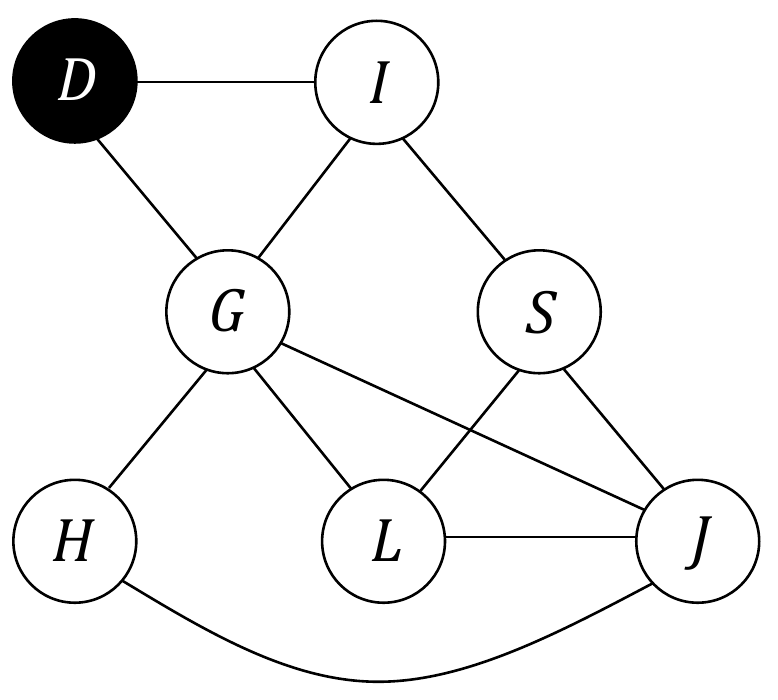} & \includegraphics[scale=0.24]{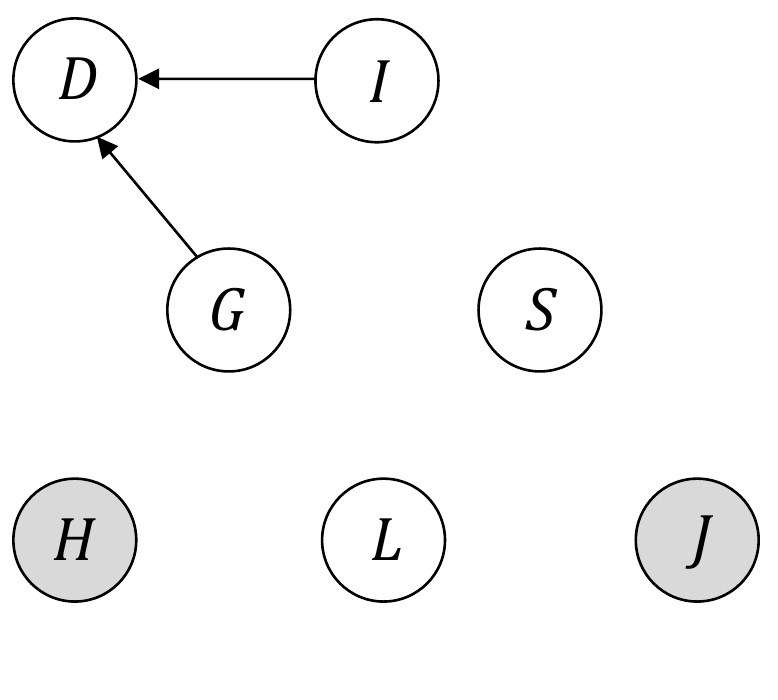} \\
	\midrule
	~2 & $I$ & $I$ & \includegraphics[scale=0.24]{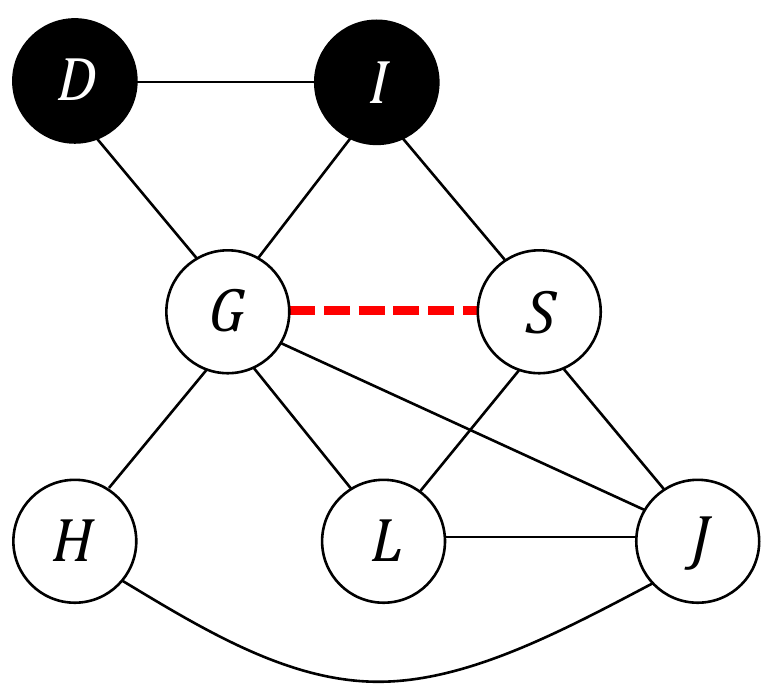} & \includegraphics[scale=0.24]{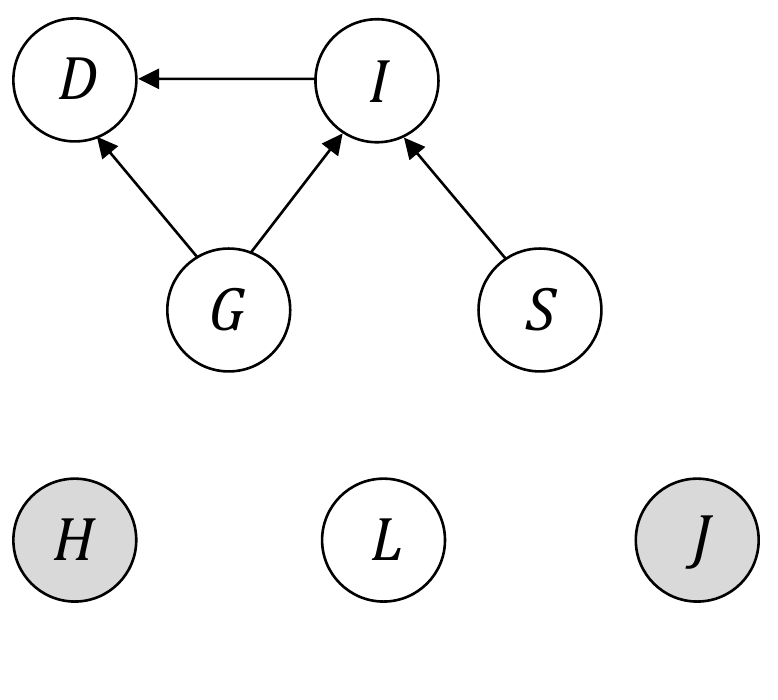} \\
	\bottomrule
	\end{tabular}
	\label{tab:inversion-example}
\end{minipage}\hspace{1.4cm}
%\hfill
\begin{minipage}{0.4\textwidth}
	\vspace{5pt}
		\centering
	\begin{tabular}{
		@{\hspace*{0ex}}m{0.14\linewidth}
		@{\hspace*{1ex}}m{0.05\linewidth}
		@{\hspace*{5ex}}m{0.05\linewidth}
		@{\hspace*{3ex}}m{0.30\linewidth}
		@{\hspace*{3ex}}m{0.30\linewidth}}
	\toprule
{\small\scshape Step}~ & ~$S$ & $v$ & ~~~~~~~$\mathcal{J}$ & ~~~~~~~~$\mathcal{H}$ \\
	\midrule
	~3 & $G,S$ & $S$ & \includegraphics[scale=0.24]{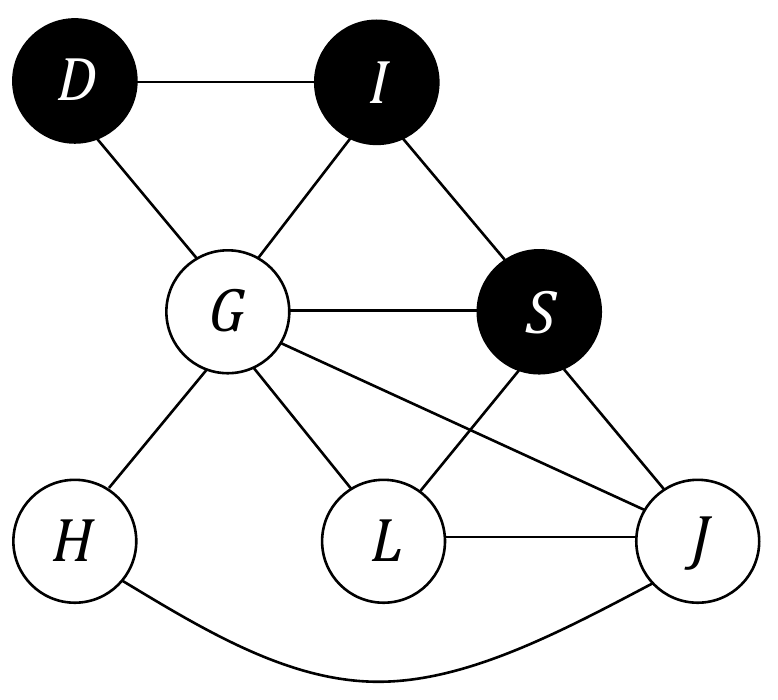} & \includegraphics[scale=0.24]{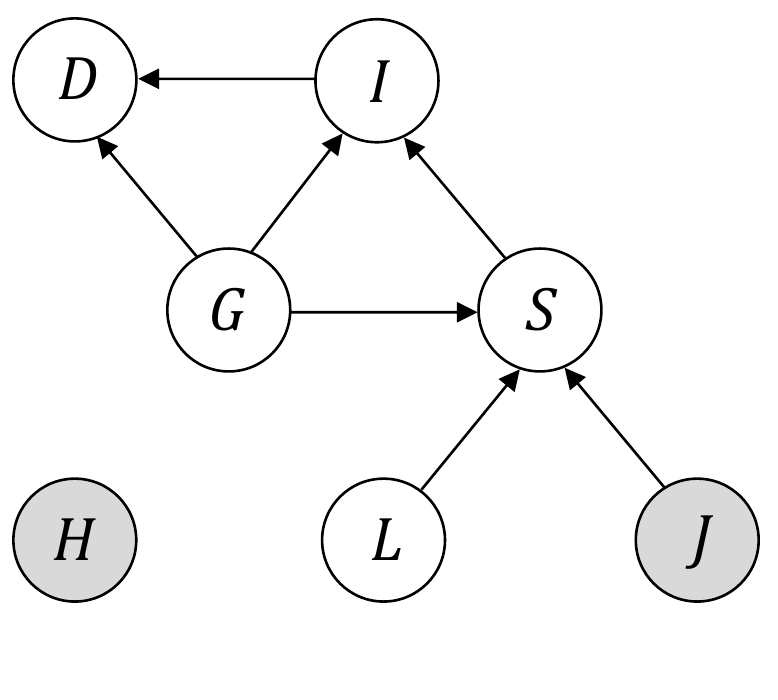} \\
	\midrule
	~4 & $G$ & $G$ & \includegraphics[scale=0.24]{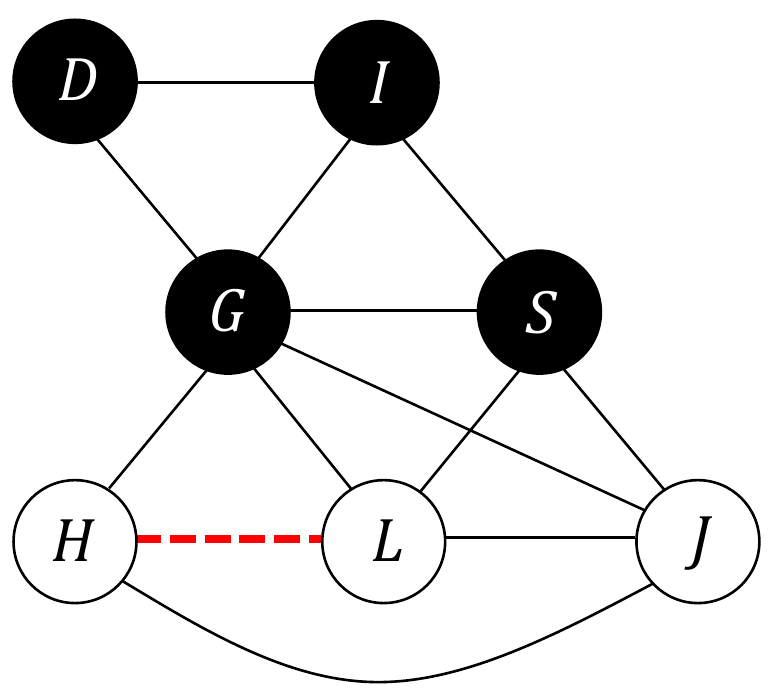} & \includegraphics[scale=0.24]{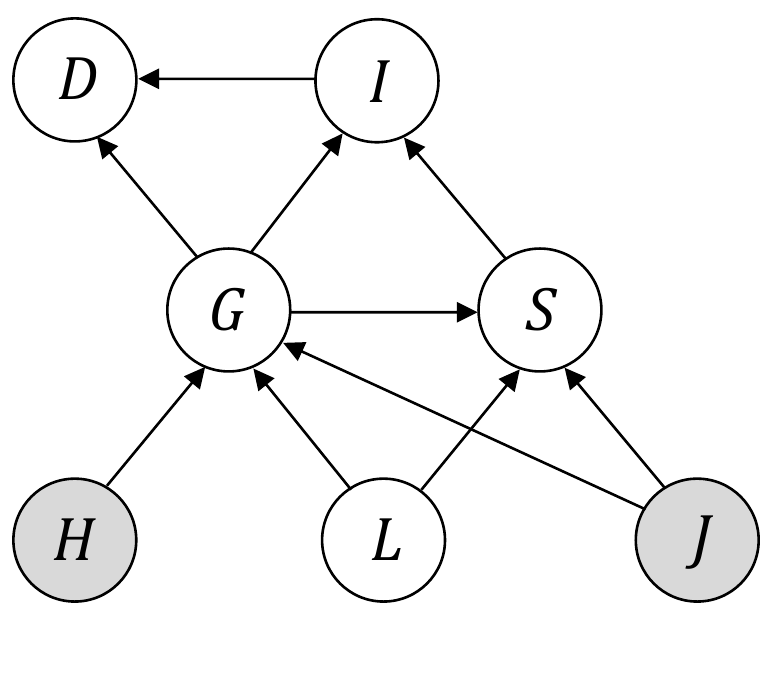} \\
	\midrule
	~5 & $L$ & $L$ & \includegraphics[scale=0.24]{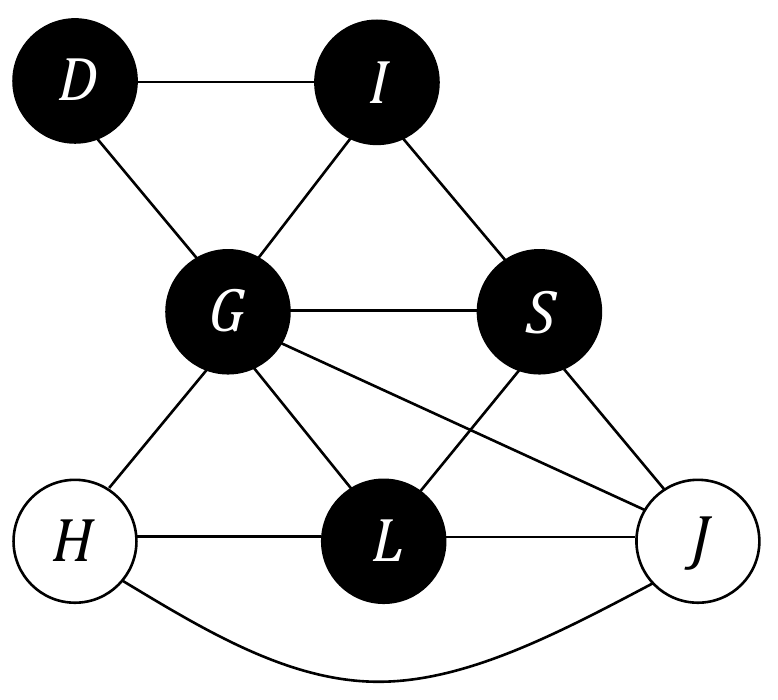} & \includegraphics[scale=0.24]{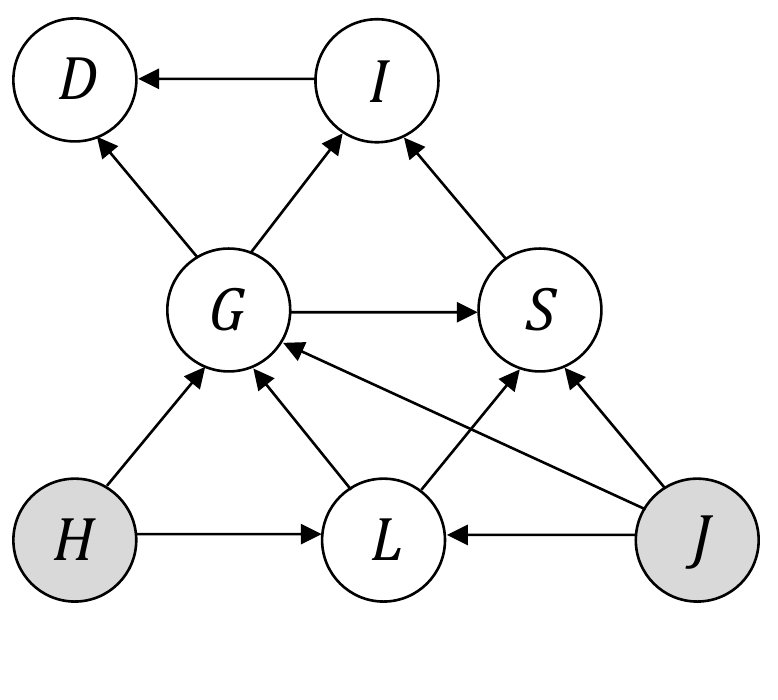} \\
	\bottomrule
	\end{tabular}
	\label{tab:inversion-example}
\end{minipage}
\vspace{-4ex}
\end{table}

%%% Local Variables:
%%% mode: latex
%%% TeX-master: "main"
%%% End:

%% file: alg-invert-bn.tex
\begin{wrapfigure}[18]{r}{0.567\textwidth}
\vspace{-2ex}	%\raggedleft
\begin{minipage}{0.555\textwidth}
\begin{algorithm}[H]
	\small
	\caption{NaMI Graph Inversion}
  \label{alg:invert-bn}
\begin{algorithmic}[1]
	\STATE {\bfseries Input:} BN structure $\mathcal{G}$, latent variables $\mathcal{Z}$, {\sc TopMode?}
	\STATE $\mathcal{J}\gets\textnormal{\sc Moralize}(\mathcal{G})$
	\STATE Set all vertices of $\mathcal{J}$ to be unmarked
	\STATE $\mathcal{H}\gets\textnormal{$\{\textsc{Variables}(\mathcal{G}),\emptyset\}$, i.e. unconnected graph}$
	\STATE $\text{\sc Upstream}\gets\text{``parent'' if {\sc TopMode?} else ``child''}$
	\STATE $\text{\sc Downstream}\gets\text{``child'' if {\sc TopMode?} else ``parent''}$
	\STATE $S\gets\textnormal{all latent variables without {\sc Upstream} latents in $\mathcal{G}$}$
		\WHILE{$S\not=\emptyset$}
			\STATE Select $v\in S$ according to min-fill criterion
			\STATE Add edges in $\mathcal{J}$ between unmarked neighbours of $v$
			\STATE Make unmarked neighbours of $v\in\mathcal{J}$, $v$'s parents in $\mathcal{H}$
			\STATE Mark $v$ and remove from $S$
			\FOR{unmarked latents {\sc Downstream} $u$ of $v$ in $\mathcal{G}$}
				\STATE \hspace*{-2ex}Add $u$ to $S$ if all its {\sc Upstream} latents in $\mathcal{G}$ are marked % $w$
			\ENDFOR
		\ENDWHILE
	\STATE \textbf{return} $\mathcal{H}$
\end{algorithmic}
\end{algorithm}
\end{minipage}%\vspace{-15pt}
\end{wrapfigure}
%%% Local Variables:
%%% mode: latex
%%% TeX-master: "main"
%%% End:

%% file: applications.tex
\section{Experiments}\label{sec:applications}
We now consider the empirical impact of using NaMI compared with previous approaches.
In \S\ref{sec:vae-experiment}, we highlight the importance of using a faithful inverse in the VAE context, demonstrating that doing so results in a tighter variational bound and a higher log-likelihood.
In \S\ref{sec:binary-tree-experiment}, we use NaMI in the fixed-model setting.
Here our results demonstrate the importance of using both a faithful and minimal inverse on the efficiency of the learned inference network.
Low-level details on the experimental setups can be found in Appendix D and an implementation at {\small \url{https://git.io/fxVQu}}.

\subsection{Relaxed Bernoulli VAEs}\label{sec:vae-experiment}
\input{fig-relaxed-bernoulli-vae}
Prior work has shown that more expressive inference networks give an improvement in amortized SVI on sigmoid belief networks and standard VAEs, relative to using the mean-field approximation \citep{UriaEtAl2016, MaaloeEtAl2016, RezendeMohamed2015, KingmaEtAl2016}. \citet{KrishnanEtAl2017} report similar results when using more expressive inverses in deep linear-chain state-space models.
It is straightforward to see that any minimally faithful inverse for the standard VAE framework~\citep{KingmaWelling2013}
has a fully connected clique over the latent variables so that the inference network can take account of the explaining-away effects between the latent variables in the generative model.
As such, both forward-NaMI and backward-NaMI produce the same inverse.

The relaxed Bernoulli VAE \citep{MaddisonEtAl2016, JangEtAl2016} is a VAE variation that replaces both the prior on the latents and the distribution over the latents given the observations with the relaxed Bernoulli distribution (also known as the Concrete distribution). It can also be understood as a ``deep'' continuous relaxation of sigmoid belief networks.

We learn a relaxed Bernoulli VAE with 30 latent variables on MNIST, comparing a faithful inference network (parametrized with MADE~\citep{GermainEtAl2015}) to the mean-field approximation, after 1000 epochs of learning for
ten different sizes of inference network, keeping the size of the generative network fixed. We note that {the mean-field inference network has the same structure as the heuristic one that reverses the edges from the generative model}. A tight bound on the marginal likelihood is estimated with annealed importance sampling (AIS) \citep{Neal1998, WuEtAl2016b}.

The results shown in Figure \ref{fig:relaxed-bernoulli-vae} indicate that using a faithful inverse on this model produces a
significant improvement in learning over the mean-field inverse.
Note that the x-axis indicates the number of parameters in the inference network. We observe that for \emph{every} capacity level, the faithful inference network has a lower negative ELBO and AIS estimate than that of the mean-field inference network. In Figure \ref{fig:vae-gap}, the variational gap is observed to decrease (or rather, the variational bound tightens) for the faithful inverse as its capacity is increased, whereas it increases for the mean-field inverse.
This example illustrates the inadequacy of the mean-field approximation in certain classes of models, in that it can result in significantly underutilizing the capacity of the model.

\subsection{Binary-tree Gaussian BNs}\label{sec:binary-tree-experiment}
\input{fig-bn-graphs}
\input{fig-binary-tree-kl}

Gaussian BNs are a class of models in which the conditional distribution of each variable is normally distributed, with a fixed variance and a mean that is a fixed linear combination of its parents plus an offset. We consider here Gaussian BNs with a binary-tree structured graph and observed leaves (see Figure \ref{fig:binary-tree-bn} for the case of depth, $d=3$). In this class of models, the exact posterior can be calculated analytically \citep[\S7.2]{KollerFriedman2009} and so it forms a convenient test-bed for performance.

The heuristic inverses simply invert the edges of the graph (Figure \ref{fig:binary-tree-heuristic}), whereas a natural minimally faithful inverse requires extra edges between subtrees (e.g. Figure \ref{fig:binary-tree-inverse}) to account for the influence one node can have on others through its parent.
For this problem, it turns out that running reverse-NaMI (Figure \ref{fig:binary-tree-most-compact}) produces
a more compact inverse than forward-NaMI. This, in fact, turns out to be the most compact possible I-map for any $d>3$.
Nonetheless, all three inversion methods have significantly fewer edges than the fully connected inverse (Figure \ref{fig:binary-tree-fully-connected}).

The model is fixed and the inference network is learnt from samples from the generative model, minimizing the ``reverse'' KL-divergence, namely that from the posterior to the inference network $\textsc{KL}(p_{\theta}(\mathbf{z} | \mathbf{x}) || q_{\psi}(\mathbf{z} | \mathbf{x}))$, as per~\citep{PaigeWood2016}.
We compared learning across the inverses produced by using Stuhlm{\"u}ller's heuristic, forward-NaMI, reverse-NaMI, and taking the fully connected inverse.
The fully connected inference network was parametrized using MADE~\citep{GermainEtAl2015}, and the forward-NaMI one with a novel MADE variant that modifies the masking matrix to exactly capture the tree-structured dependencies (see Appendix E.2). As the same MADE approaches cannot be used for heuristic and reverse-NaMI inference networks, these were instead parametrized with a separate neural network for each variable's density function.
The inference network sizes were kept constant across approaches.

Results are given in Figure \ref{fig:binary-tree-kl} for depth $d=5$ averaging over 10 runs. Figures \ref{fig:binary-tree-kl-train} and \ref{fig:binary-tree-kl-test} show an estimate of $\textsc{KL}(p_{\theta}(\mathbf{z} | \mathbf{x}) || q_{\psi}(\mathbf{z} | \mathbf{x}))$ using the train and test sets respectively. From this, we observe that it is necessary to model at least the edges in an I-map for the inference network to be able to recover the posterior, and convergence is faster with fewer edges in the inference network.\input{fig-gmm}
Despite the more compact reverse-NaMI inverse converging faster than the forward-NaMI one, the latter seems to converges to a better final solution.  This may be because the MADE approach could not be used for the reverse-NaMI inverse, but this is a subject for future investigation nonetheless.

Figure \ref{fig:binary-tree-kl-posterior} shows the average negative log-likelihood of 200 samples from the inference networks evaluated on the analytical posterior, conditioning on five fixed datasets sampled from the generative model not seen during learning. It is thus a measure of how successful inference amortiziation has been. All three faithful inference networks have significantly lower variance over runs compared to the unfaithful inference network produced by Stuhlm{\"u}ller's algorithm.

We also observed during other experimentation that if one were to decrease the capacity of all methods, learning remains stable in the natural minimally faithful inverse at a threshold where it becomes unstable in the fully connected case and in Stuhlm{\"u}ller's inverse.

\subsection{Gaussian Mixture Models}\label{sec:gmm-experiment}

Gaussian mixture models (GMMs) are a clustering model where the data
$\mathbf{x}=\{x_1,x_2,\ldots,x_N\}$ is
assumed to have been generated from one of $K$ clusters, each of which
has a Gaussian distribution with parameters $\{\mu_j,\Sigma_j\}$, $j=1,2,\ldots,K$.   Each datum, $x_i$ is associated with a corresponding index, $z_i\in\{1,\ldots,K\}$ that gives the identity of that datum's cluster. The indices, $\mathbf{z}'=\{z_i\}$ are drawn i.i.d. from a categorical distribution with parameter $\phi$. Prior distributions are placed on $\theta=\{\mu_1,\Sigma_1,\ldots,\mu_K,\Sigma_K\}$ and $\phi$, so that the latent variables are $\mathbf{z}=\{\mathbf{z}',\theta,\phi\}$. The goal of inference is then to determine the posterior $p(\mathbf{z}\mid\mathbf{x})$, or some statistic of it.

As per the previous experiment, this falls into the fixed-model setting.
We factor the fully-connected inverse as, $q(\theta|x)q(\phi|\theta,\mathbf{x})q(\mathbf{z}'|\phi,\theta,\mathbf{x})$.
It turns out that applying reverse-NaMI decouples the dependence between the indices, $\mathbf{z}'$, and produces a much more compact factorization, $q(\theta|\mathbf{x},\phi)\prod^N_iq(z_i|x_i,\phi,\theta)q(\phi|\mathbf{x})$, than either the fully-connected or forward-NaMI inverses for this model. The inverse structure produced by Stuhlm\"uller's heuristic algorithm is very similar to the reverse-NaMI structure for this problem and is omitted.

We train our amortization artifact over datasets with $N=200$ samples and $K=3$ clusters. The inference network terms with distributions over vectors were parametrized by MADE, and we compare the results for the fully-connected and reverse-NaMI inverses.  We hold the neural network capacities constant across methods and average over 10 runs, the results for which are shown in Figure \ref{fig:gmm}. We see that learning is faster for the minimally faithful reverse-NaMI method, relative to the fully-connected inverse, and converges to a better solution, in agreement with the other experiments.

\subsection{Minimal and Non-minimal Faithful Inverses}\label{sec:skip-experiment}
\begin{wrapfigure}{r}{0.5\textwidth}
	\vspace{-8pt}
	\centering
	\subcaptionbox{12 skips edges\label{fig:skip1}}
	{\includegraphics[width=0.45\linewidth]{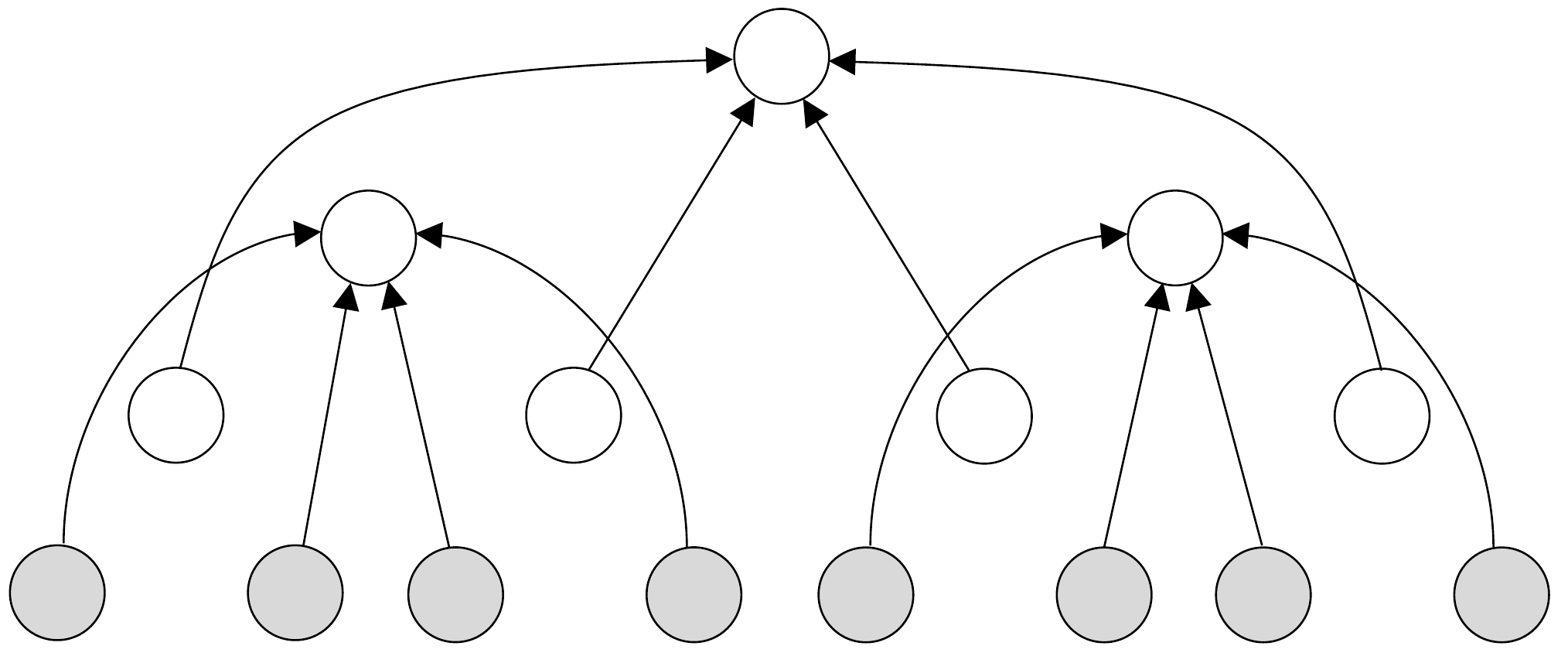}}\hspace{0.5cm}
	\subcaptionbox{16 skips edges\label{fig:skip2}}
	{\includegraphics[width=0.45\linewidth]{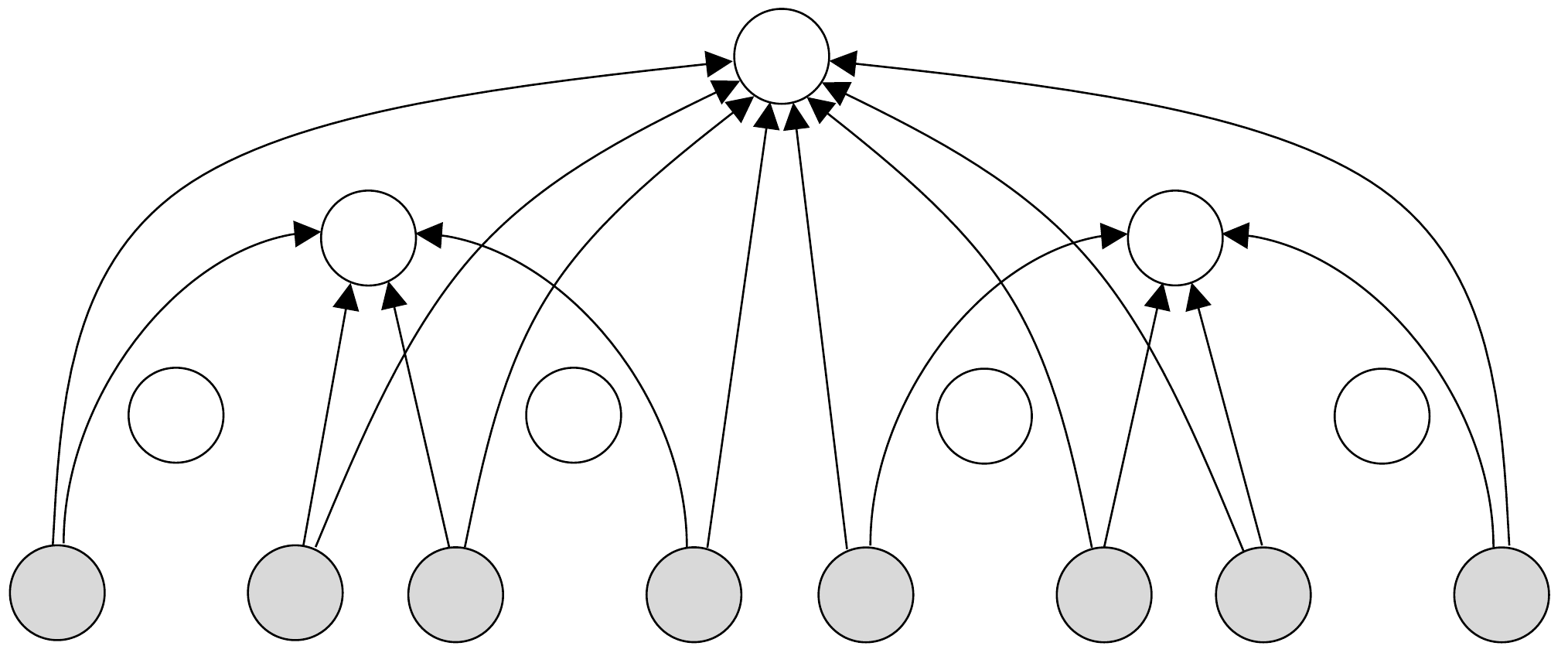}}
	\caption{Additional edges over forward-NaMI. \label{fig:skip-model}}\vspace{-10pt}
\end{wrapfigure}
To further examine the hypothesis that a non-minimal
faithful inverse has slower learning and converges to a worse
solution relative to a minimal one, we performed the setup of Experiment \ref{sec:binary-tree-experiment} with depth d = 4,  comparing the forward-NaMI
network to two additional networks that added 12 and 16 connections to forward-NaMI (holding the total capacity fixed).
\begin{wrapfigure}{r}{0.4\textwidth}
	\vspace{-15pt}
  \centering
  {\includegraphics[width=\linewidth,trim={0 0 0 0.6cm},clip]{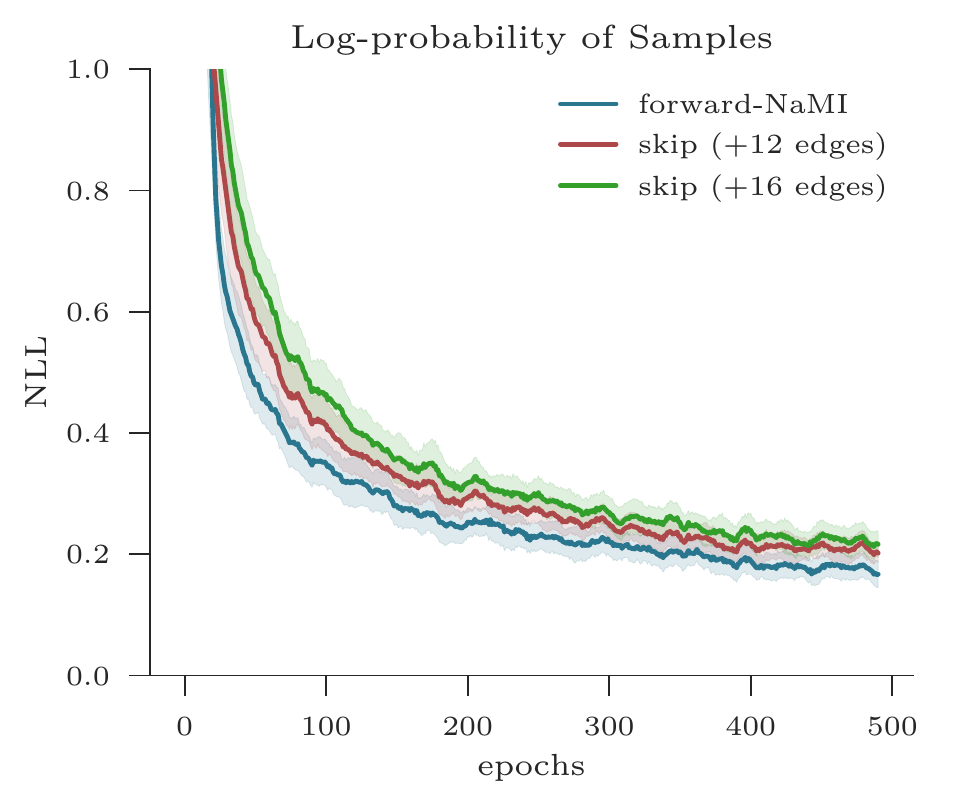}}
  \vspace*{-3ex}
		\caption{Average NLL of inference network samples under analytical posterior. \label{fig:exp5-results}}
\end{wrapfigure}

\vspace{-2pt}
The additional edges are shown  in Figure \ref{fig:skip-model}.
Note the
regular forward-NaMI edges are omitted for visual clarity.

Figure \ref{fig:exp5-results} shows the average negative log likelihood (NLL) under the true posterior for samples generated by the inference network, based on 5 datasets not seen during training. It appears that the more edges are added beyond minimality, the slower is the initial learning and convergence is to a worse solution.

To further explain why minimality is crucial, we note that adding additional edges beyond minimality means that there will be factors that condition on variables whose probabilistic influence is blocked by the other variables. This effectively adds an input of random noise into these factors, which is why we then see slower learning and convergence to a worse solution.

%%% Local Variables:
%%% mode: latex
%%% TeX-master: "main"
%%% End:

%% file: fig-relaxed-bernoulli-vae.tex
\begin{figure}[t]
  \centering
  \subcaptionbox{}%
  {\includegraphics[width=0.32\textwidth]{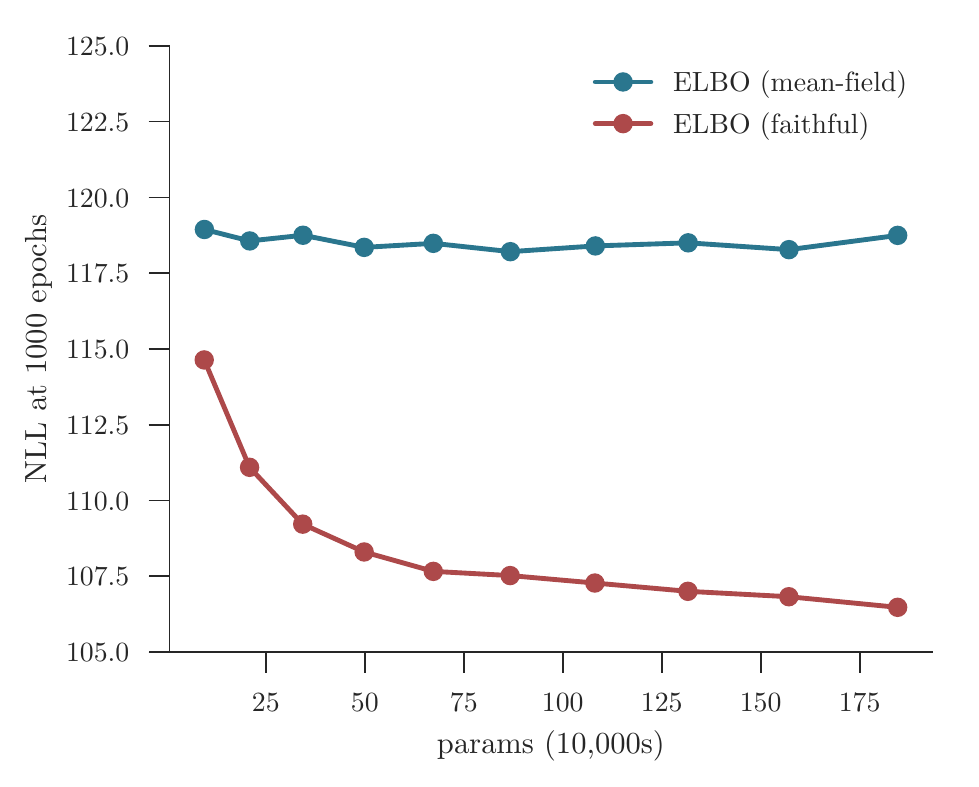}}
  \,
  \subcaptionbox{}%
  {\includegraphics[width=0.32\textwidth]{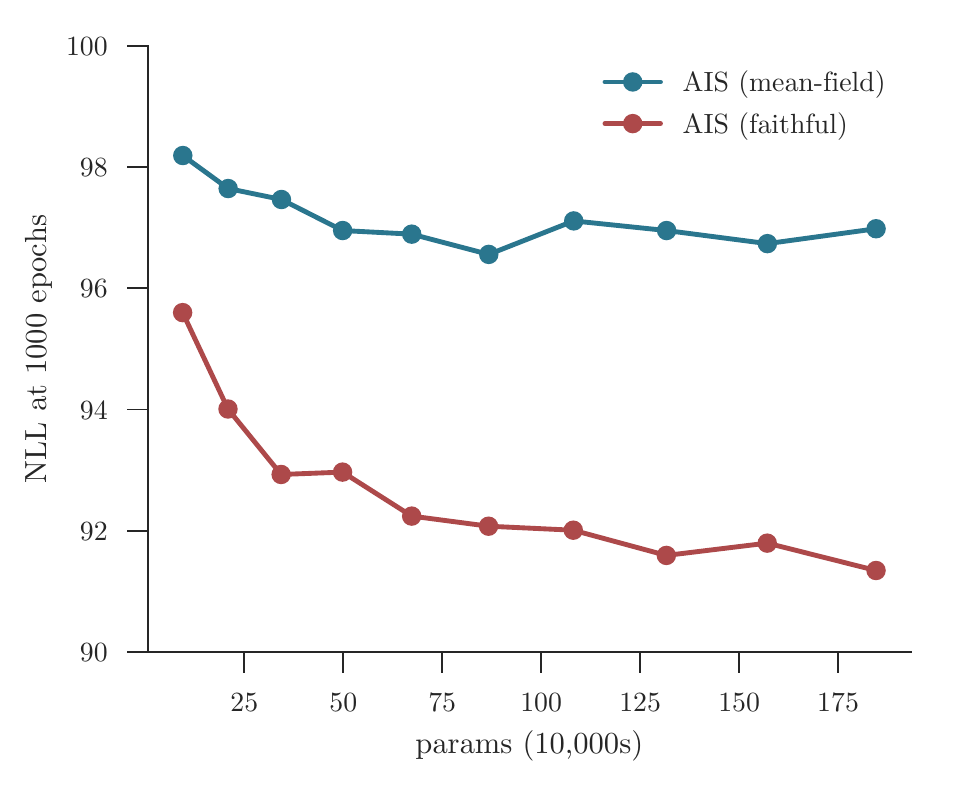}}
  \,
  \subcaptionbox{\label{fig:vae-gap}}%
  {\includegraphics[width=0.32\textwidth]{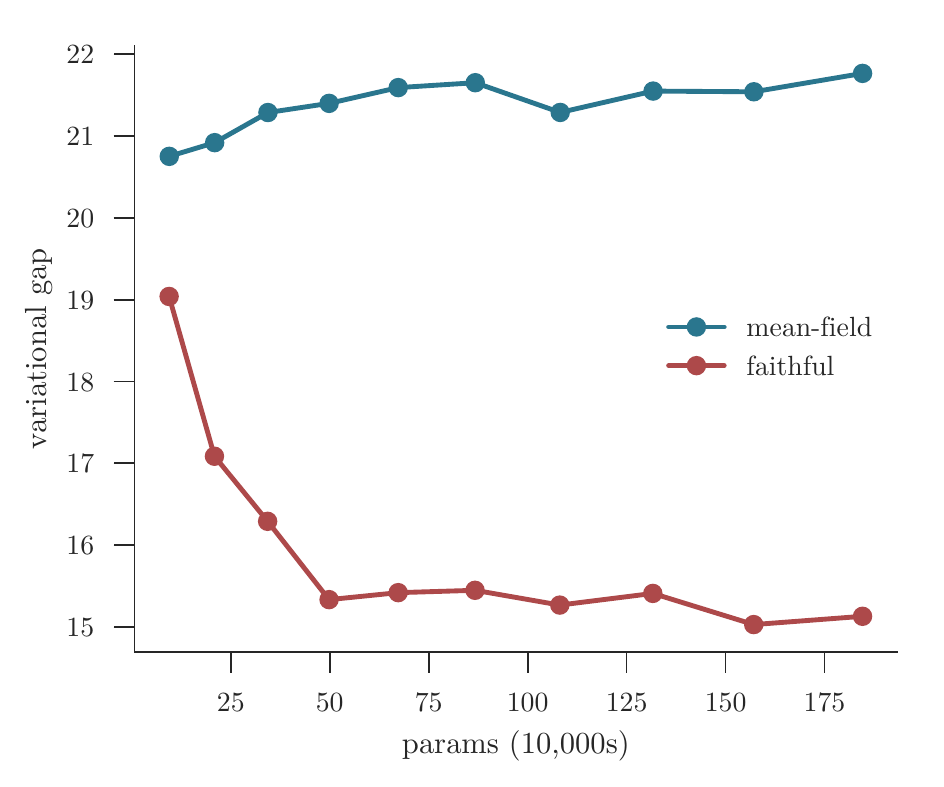}}
  \caption[Results for relaxed Bernoulli VAEs]{Results for the relaxed Bernoulli VAE with 30 latent units, compared after 1000 epochs of learning the: (a) negative ELBO, and (b) negative AIS estimates, varying inference network factorizations and capacities (total number of parameters); (c) An estimate of the variational gap, that is, the difference between marginal log-likelihood and the ELBO.}
  \vspace*{-2ex}
  \label{fig:relaxed-bernoulli-vae}
\end{figure}

%% file: fig-bn-graphs.tex
\begin{figure}[t]
  \centering
  \subcaptionbox{\label{fig:binary-tree-bn}}%
  {\includegraphics[width=0.19\textwidth]{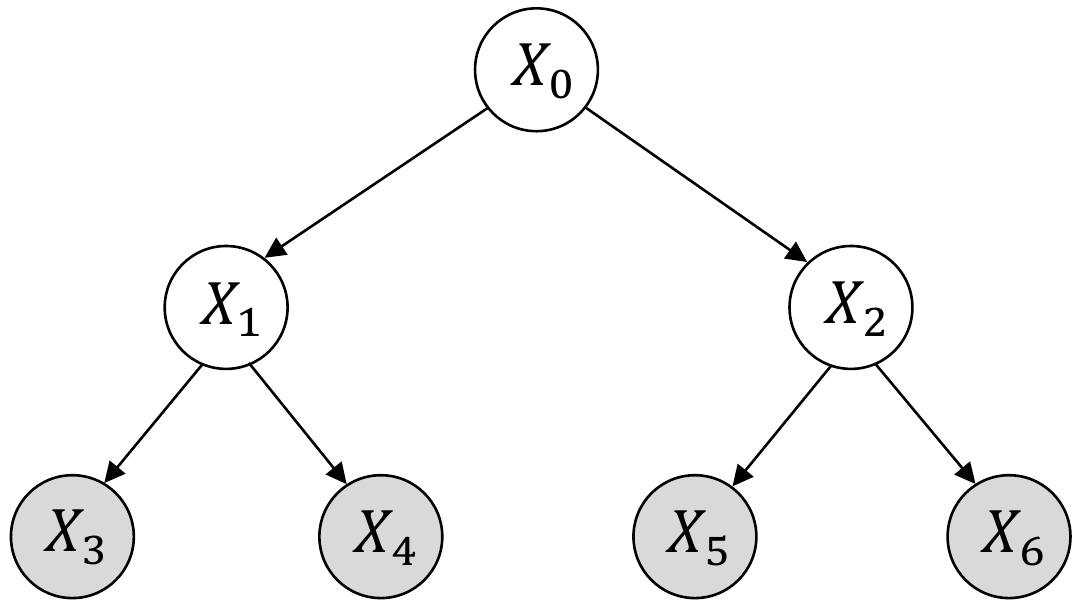}}%
  \,\,%
  \subcaptionbox{\label{fig:binary-tree-heuristic}}%
  {\includegraphics[width=0.19\textwidth]{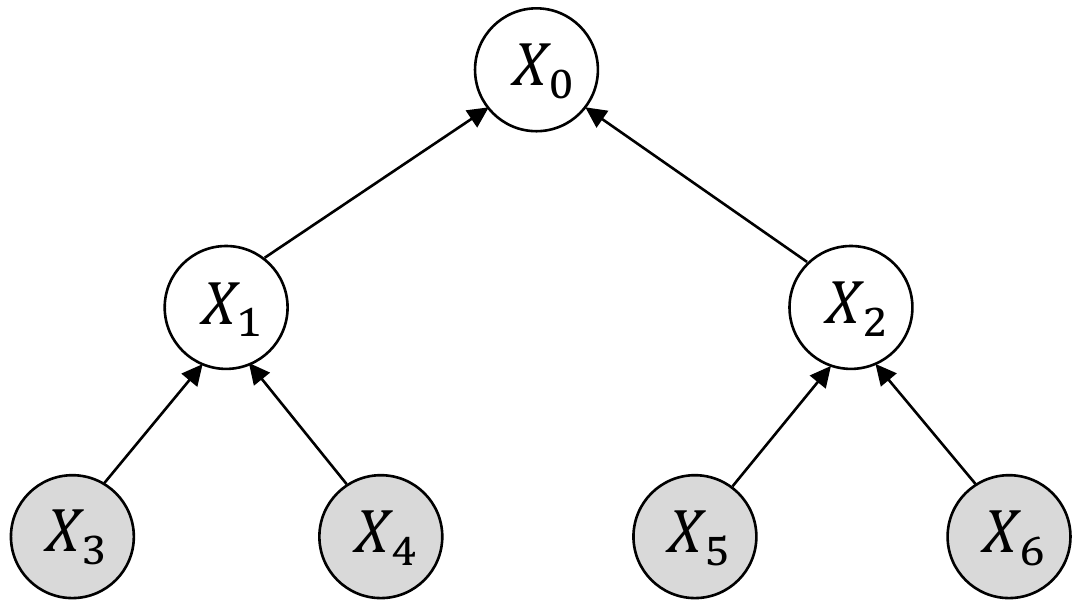}}%
  \,\,%
  \subcaptionbox{\label{fig:binary-tree-inverse}}%
  {\includegraphics[width=0.19\textwidth]{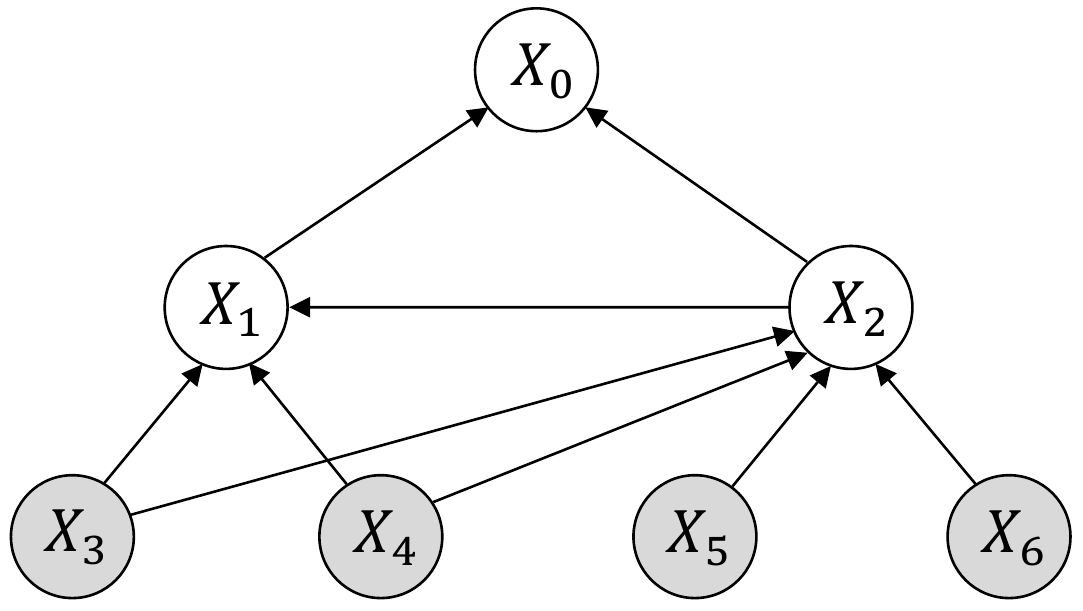}}%
  \,\,%
  \subcaptionbox{\label{fig:binary-tree-most-compact}}%
  {\includegraphics[width=0.19\textwidth]{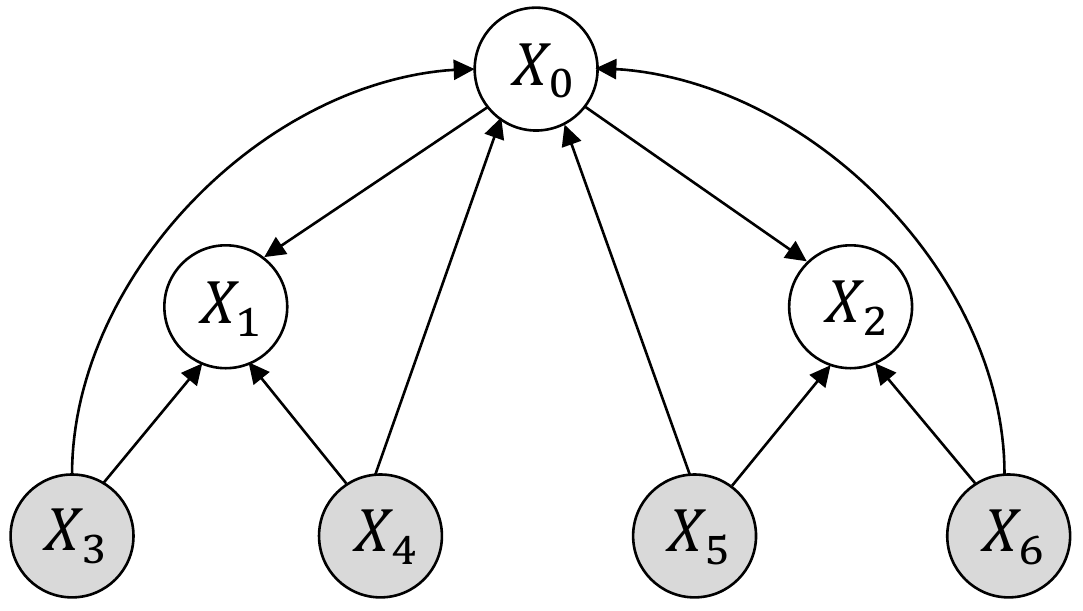}}%
  \,\,%
  \subcaptionbox{\label{fig:binary-tree-fully-connected}}%
  {\includegraphics[width=0.19\textwidth]{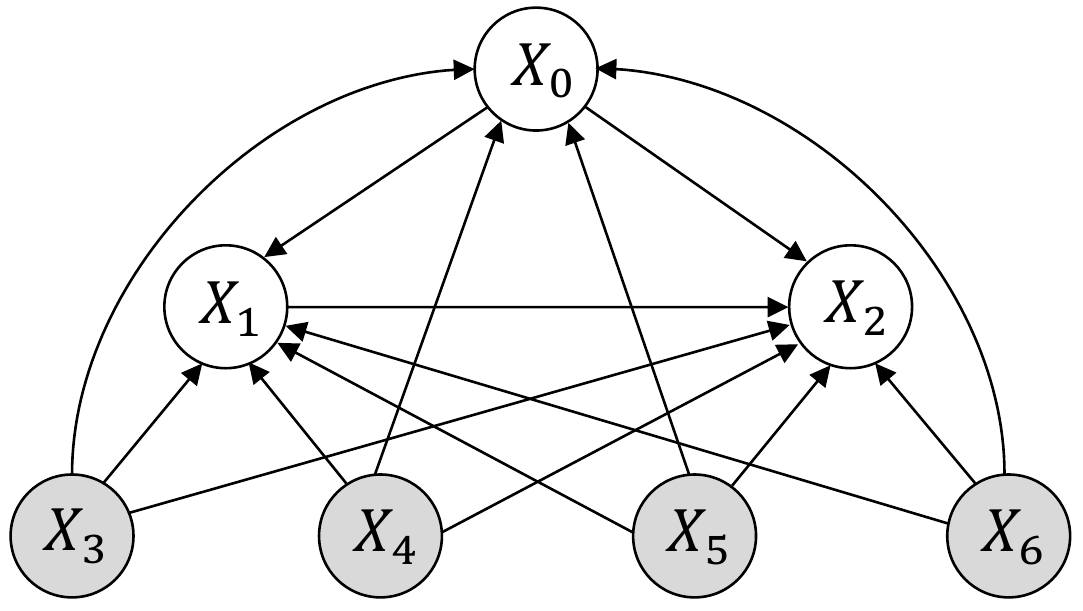}}%
	\caption{(a) BN structure for a binary tree with $d=3$; (b) Stuhlm{\"u}ller's heuristic inverse; (c) Natural minimally faithful inverse produced by NaMI in topological mode; (d) Most compact inverse when $d>3$, given by running NaMI in reverse topological mode; (e) Fully connected inverse.}
 % \vspace{-1ex}
  \label{fig:bn-graphs}
\end{figure}

%% file: fig-binary-tree-kl.tex
\begin{figure}[t]
  \centering
  \subcaptionbox{\label{fig:binary-tree-kl-train}}%
  {\includegraphics[width=0.33\textwidth]{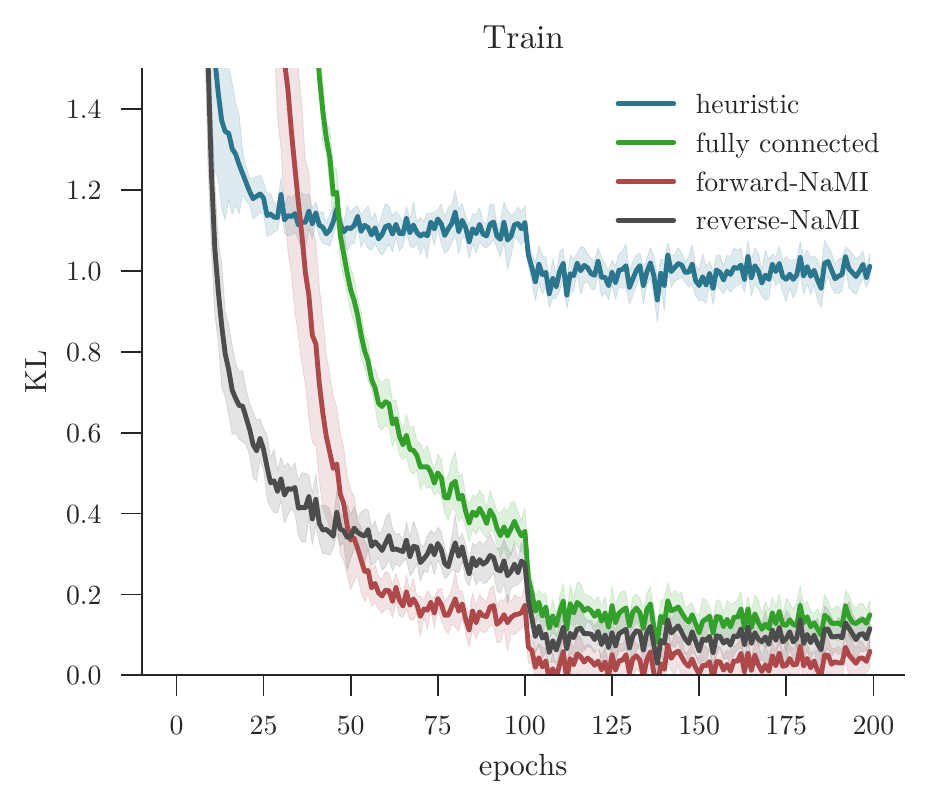}}%
  \,%
  \subcaptionbox{\label{fig:binary-tree-kl-test}}%
  {\includegraphics[width=0.33\textwidth]{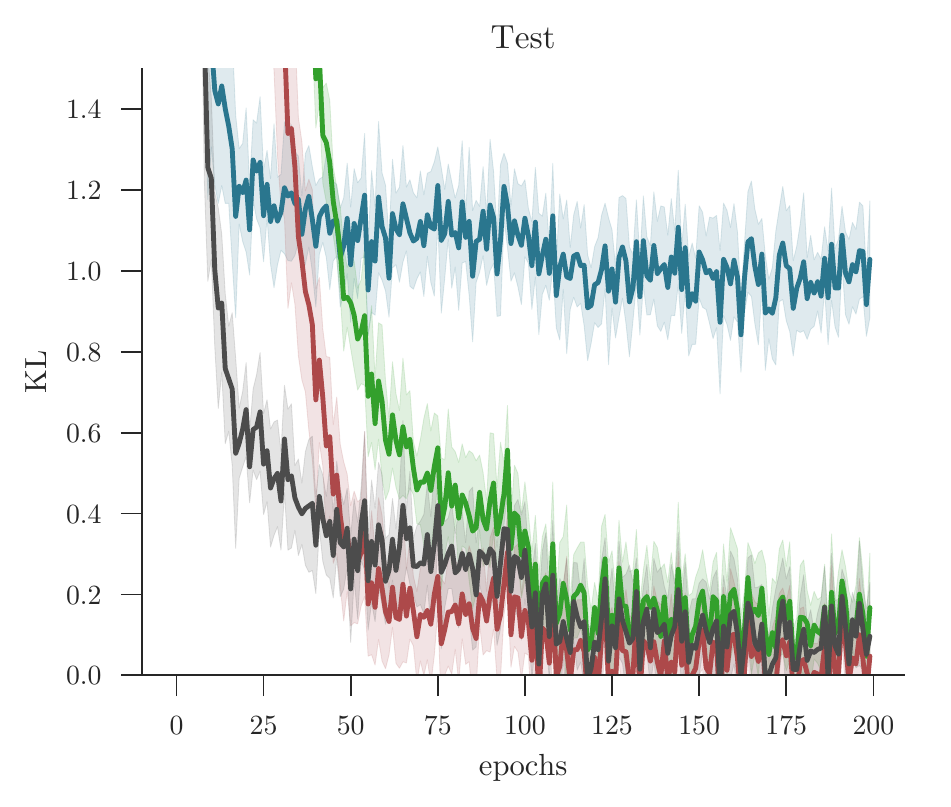}}%
  \,%
  \subcaptionbox{\label{fig:binary-tree-kl-posterior}}%
  {\includegraphics[width=0.33\textwidth]{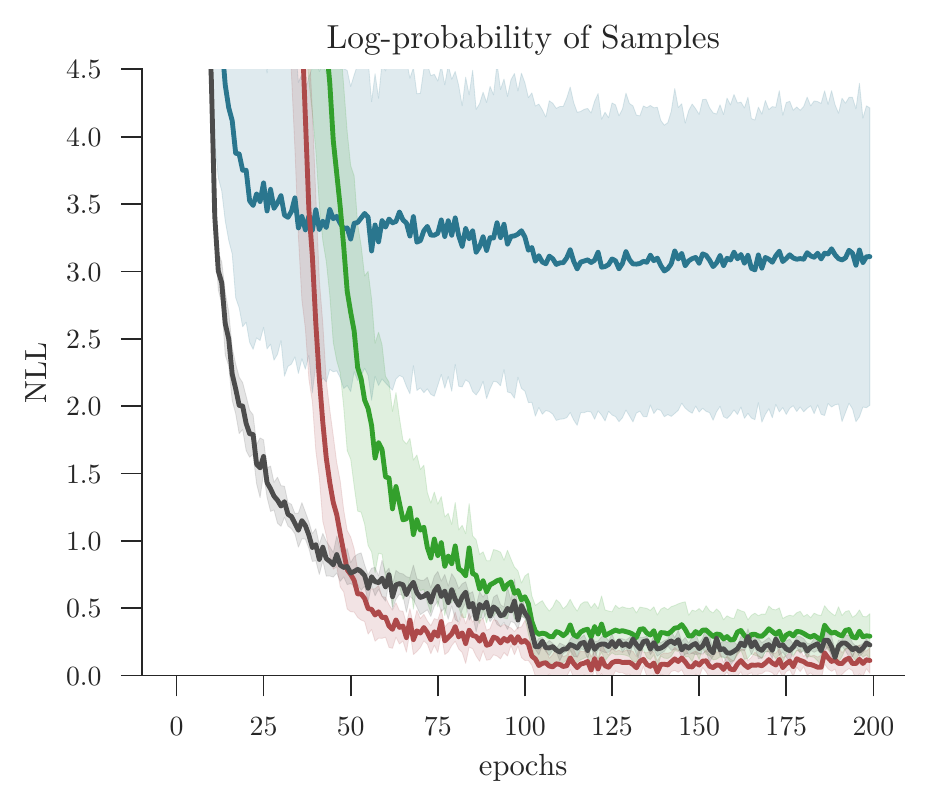}}%
  \vspace{-1ex}
  \caption[Results for binary trees]{
    Results for binary tree Gaussian BNs with depth $d=5$, comparing inference network factorizations in the compiled inference setting.
    The KL divergence from the analytical posterior estimated to the inference network on the training and test sets are shown
    in (a) and (b) respectively.
    (c) shows the average negative log-likelihood of inference network samples under the analytical posterior, conditioning on five held-out data sets.
    The results are averaged over 10 runs and 0.75 standard deviations indicated.
		The drop at 100 epochs is due to decimating the learning rate.}
 \vspace{-2.5ex}
  \label{fig:binary-tree-kl}
\end{figure}

%% file: fig-gmm.tex
\begin{figure}[t]
  \centering
  \subcaptionbox{\label{fig:gmm-train}}%
  {\includegraphics[width=0.4\textwidth]{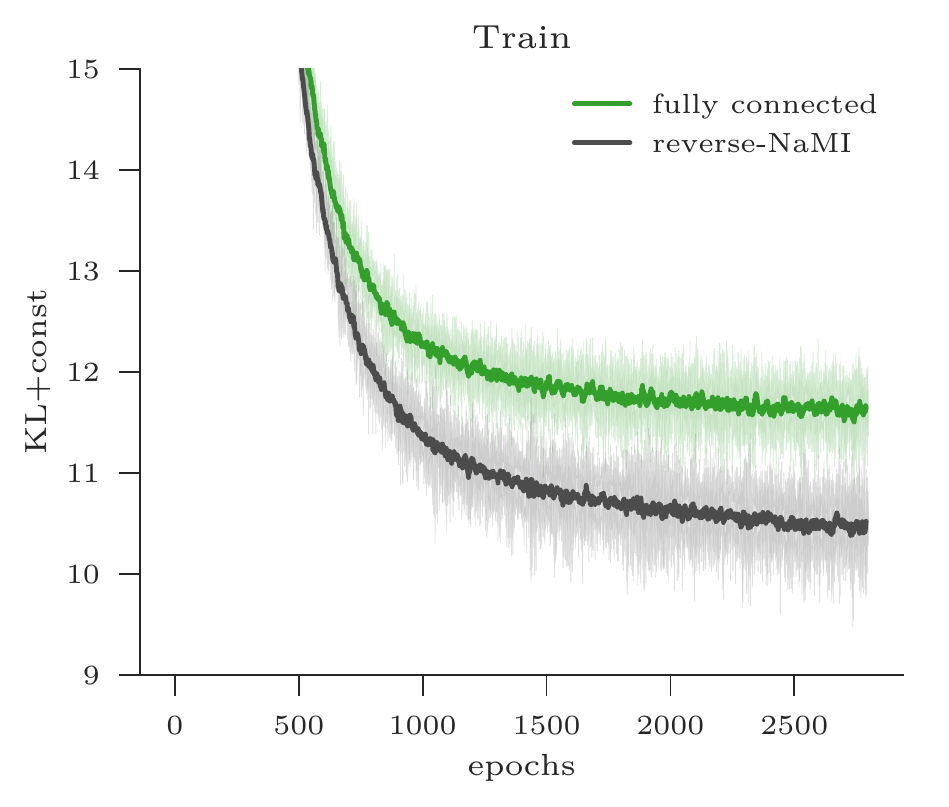}}
  ~~~~~~~~~~
  \subcaptionbox{\label{fig:gmm-test}}% 
  {\includegraphics[width=0.4\textwidth]{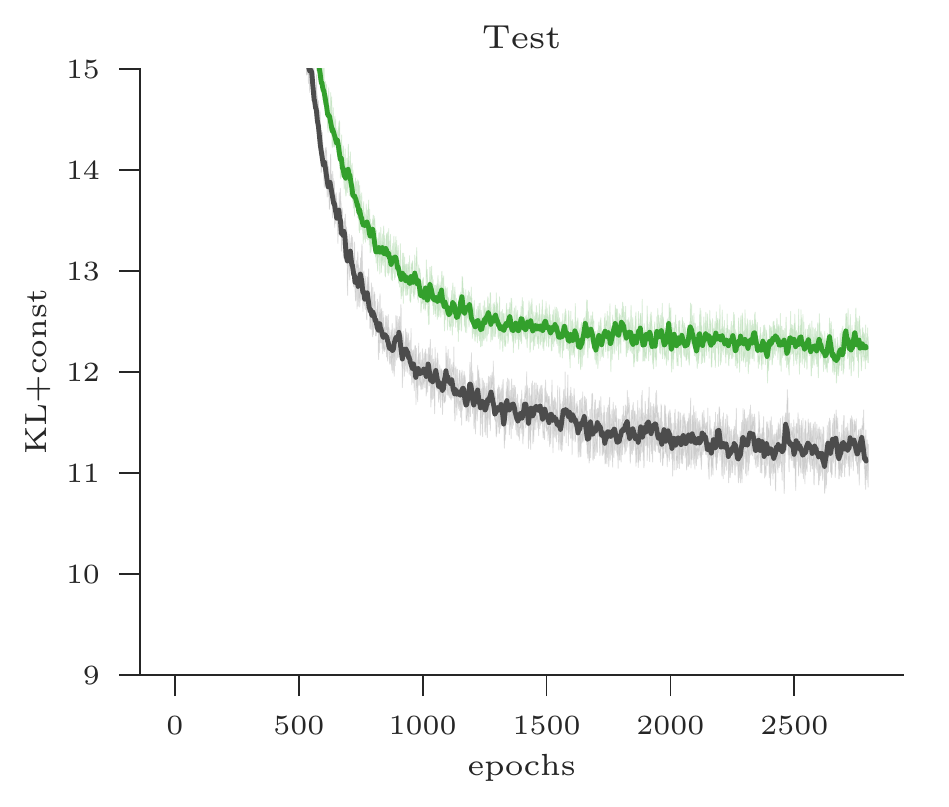}}
  \vspace*{-0.4ex}
	\caption[Simple examples]{Convergence of reverse KL divergence (used as the training objective) for Bayesian GMM for $K=3$ clusters and $N=200$ data points, comparing inference networks with a fixed generative model. The shaded regions indicate 1 standard error in the estimation.
\vspace*{-2.2ex}	}
	\label{fig:gmm}
\end{figure}

%% file: discussion.tex
\section{Discussion}
We have presented NaMI, a tractable framework that, given the BN structure for a generative model, produces a natural factorization for its inverse that is a minimal I-map for the model.  We have argued that this should be used to guide the design of the coarse-grain structure of the inference network in amortized inference. Having empirically analyzed the implications of using NaMI, we find that it learns better inference networks than previous heuristic approaches.
We further found that, in the context of VAEs, improved inference networks have a knock-on effect on the generative network, improving the generative networks as well.

Our framework opens new possibilities for learning structured deep generative models that combine traditional Bayesian modeling by probabilistic graphical models with deep neural networks. This allows us to leverage our typically strong knowledge of which variables effect which others, while not overly relying on our weak knowledge of the exact functional form these relationships take.

To see this, note that if we forgo the niceties of making mean-field assumptions,
we can impose arbitrary structure on a generative model simply by
controlling its parameterization.  The only requirement on the generative network
to evaluate the ELBO is that
we can evaluate the network density at a given input.  Recent
advances in normalizing flows \citep{HuangEtAl2018, ChenEtAl2018} mean it is possible
to construct flexible and general purpose distributions that satisfy this requirement
and are amenable to application of dependency constraints from our
graphical model.  This obviates the need to make assumptions such
as conjugacy as done by, for example,~\citet{JohnsonEtAl2016}.

NaMI provides a critical component to constructing such a framework, as it allows one to ensure that the inference
network respects the structural assumptions imposed on the generative network,
without which a tight variational bound cannot be achieved.

%%% Local Variables:
%%% mode: latex
%%% TeX-master: "main"
%%% End:

%% file: acknowledgements.tex
\section*{Acknowledgments}
We would like to thank (in alphabetical order) Rob Cornish, Rahul Krishnan, Brooks Paige, and Hongseok Yang for their thoughtful help and suggestions.

SW and AG gratefully acknowledge support from the EPSRC AIMS CDT
through grant EP/L015987/2. RZ acknowledges support under DARPA D3M,
under Cooperative Agreement FA8750-17-2-0093. NS was supported by
EPSRC/MURI grant EP/N019474/1. TR and YWT are supported in part by the
European Research Council under the European Union’s Seventh Framework
Programme (FP7/2007–2013) / ERC grant agreement no. 617071. 
TR further acknowledges support of the ERC StG IDIU. FW was
supported by The Alan Turing Institute under the EPSRC grant
EP/N510129/1, DARPA PPAML through the U.S. AFRL under Cooperative
Agreement FA8750-14-2-0006, an Intel Big Data Center grant, and DARPA
D3M, under Cooperative Agreement FA8750-17-2-0093.

%% file: background.tex
% !TEX root = supplementary.tex

\section{Theory}\label{sec:theory}
Here, we examine the complexity of the inversion problem and prove the correctness of NaMI's graph inversion.

\subsection{Inversion complexity}\label{sec:problem-hardness}
To understand the theoretical gains we obtain, it is useful to compare it with a simpler, but suboptimal, alternate that uses the d-separation properties of a BN structure to form a minimally faithful inverse.
By the general product rule, any distribution over $\mathbf{y}=\{y_1,\ldots,y_n\}$ can be factored as
$p(\mathbf{y}) = \prod_{i=1}^np(y_i\mid y_1,\ldots,y_{i-1})$,
for any ordering of $\mathbf{y}$.
The conditioning sets, $\{y_1,\ldots,y_{i-1}\}$, can be restricted according to the conditional independence assertions made by $p$.
To produce a minimal I-map, they can be restricted as
$p(\mathbf{y}) = \prod_{i=1}^np(y_i\mid \tilde{\mathbf{y}_i}\subseteq\{y_1,\ldots,y_{i-1}\})$
where $\tilde{\mathbf{y}_i}$ is a minimal subset such that $y_i\perp(\{y_1,\ldots,y_{i-1}\}\setminus\tilde{\mathbf{y}_i})\mid\tilde{\mathbf{y}_i}$.

Consequently, one could instead produce a minimally faithful inverse for $p(\mathbf{z}\mid\mathbf{x})p(\mathbf{x})$ as follows.
Set $\mathbf{y}=(\mathbf{z},\mathbf{x})$ to have an arbitrary topological ordering on $\mathbf{z}$ and $\mathbf{x}$, separately.
Initialize $\tilde{\mathbf{y}_i}=\{y_1,\ldots,y_{i-1}\}$.
Scan through $y_j\in\tilde{\mathbf{y}_i}$, removing each one if $y_i\perp y_j\mid\tilde{\mathbf{y}_i}\setminus\{y_j\}$, repeating until none can be removed and $\tilde{\mathbf{y}_i}$ is a minimal subset.

In the worst case for this naive approach, we must scan through $O(n^2)$ variables $n$ times, and the cost of determining whether to remove a variable from $\tilde{\mathbf{y}_i}$ is $O(n)$ \citep[Algorithm 3.1]{KollerFriedman2009}.
Thus, this naive method has running time $O(n^4)$.
NaMI's graph reversal, in contrast has a running time of order $O(nc)$ where $n$ is the number of variables in the graph and $c<<n$ is the size of the largest clique in the induced graph.

\subsection{Proof of correctness}\vspace{3pt}
\begin{theorem}\label{theorem:correctness}
	The Natural Minimal I-Map Generator of Algorithm 1 produces inverse factorizations that are natural and minimally faithful.
\end{theorem}\vspace{-11pt}
\begin{proof}

As in the main paper, let $\tau$ denote the reverse of the order in which variables were selected for elimination
such that $\tau$ is a permutation of $1,\dots,n$ and $\tau(n)$ is the first variable eliminated.  

We first show that inverse structure $\mathcal{H}$ produced by Algorithm 1 is guaranteed to be a valid inverse factorization, that is, it factors as
\begin{align}
\label{eq:H-fact}
q_\psi(\mathbf{z}\mid\mathbf{x})=\prod^n_{i=1} q_{i}(z_{\tau(i)}\mid\text{Pa}_\mathcal{H}(z_{\tau(i)}))
=\prod^n_{i=1} q_{i}(z_{\tau(i)}\mid\text{Pa}_\mathcal{H}(z_{\tau(i)}),\mathbf{x})
\end{align}
where $\mathbf{z}$ and $\mathbf{x}$ are the observed and latent variables, and
$\text{Pa}_\mathcal{H}(z_{\tau(i)})\subseteq\left\{\mathbf{x},z_{\tau(1)},\dots,z_{\tau(i-1)}\right\}$ indicates the parents of $z_{\tau(i)}$ in $\mathcal{H}$.
There are two critical features this factorization encapsulates that we need to demonstrate to show
  $\mathcal{H}$ provides a valid  inverse factorization: $\mathbf{x}$ only appears in the conditioning variables (i.e. 
  there are no density terms over observations) and all terms can, if desired, be conditioned on the full
  set of observations.  
  
  The former of these straightforwardly always holds, since we only add edges \emph{into} latent variables when the inverse, $\mathcal{H}$, is constructed (Line 11 in Algorithm 1). Therefore, the algorithm can never add in edges \emph{to} an observed node.  
  
  The latter is more subtle, as NaMI may produce factors which are not explicitly
  conditioned on all the observations.  However, because, as we demonstrate later, the inversion is faithful, the
  corresponding $z_{\tau(i)}$ must be conditionally independent of the all observations which are not parent nodes, given
  the state of the parent nodes.  In other words, if the inversion is faithful, this ensures that 
  each $q_{i}(z_{\tau(i)}\mid\text{Pa}_\mathcal{H}(z_{\tau(i)}))=
  q_{i}(z_{\tau(i)}\mid\text{Pa}_\mathcal{H}(z_{\tau(i)}),\mathbf{x})$, and
  thus that we have a valid inverse factorization.

Next, we prove that Algorithm 1 produces natural inverses. A general observation is that if $z_i$ is eliminated after $z_j$, there cannot be an edge from $z_j$ to $z_i$ in $\mathcal{H}$.
When the algorithm is run in topological mode, variable elimination is simulated in a topological ordering, and so all of a variable's descendants are eliminated after it is.
Therefore there cannot be an edge from a variable to its descendant, and hence the factorization is natural. An equivalent argument applies when the algorithm is run in the reverse topological mode.

Finally, we prove that the inverse factorization is minimally faithful. At a high-level, our proof consists of
showing an equivalence to a process where we
start with a fully connected graph over the variables and sequentially prune edges in the graph according to
the independencies revealed by the clique tree generated from simulating variable elimination, 
terminating when no more edges can be pruned.  
By showing that
each individual pruning never induces an unfaithful independence, we are able to demonstrate that the graphs at
each iteration of this process---including the final inverse graph---is faithful, while minimality follows from 
the fact that the process terminates when it is not possible to prune any given edge.

More precisely, by the general product rule, 
$p(\mathbf{z}|\mathbf{x})=\prod^n_{i=1}p(z_{\tau(i)}|z_{\tau(<i)},\mathbf{x})$,
where $z_{\tau(<i)}=\{z_{\tau(1)},\ldots,z_{\tau(i-1)}\}$,
for any possible $\tau$,
and any graph with this factorization is an I-map for the posterior. Each term can be simplified according the conditional independencies encoded by the posterior and the corresponding graph will still be an I-map for the posterior. For instance, if $z_{\tau(i)}$ is independent of $\{z_{\tau(1)}, \ldots,z_{\tau(i-2)}\}$ given $\{z_{\tau(i-1)},\mathbf{x}\}$, then $p(z_{\tau(i)}|z_{\tau(<i)},\mathbf{x})=p(z_{\tau(i)}|z_{\tau(i-1)},\mathbf{x})$. %, which corresponds to removing the 
By definition, the variable elimination is run in the opposite order to $\tau$.
This produces a unique corresponding
clique tree (see Appendix A.5). Furthermore, because we introduce a new factor at each iteration, the
number of cliques in this clique tree matches the number of latent variables in the original BN, with each
clique being associated with the corresponding variable that was eliminated at that iteration (though the
clique itself may contain multiple variables).  We can thus define $C_{\tau(i)}$ as the unique clique
corresponding to the elimination of $z_{(\tau(i))}$.  Further, we can define $S_{\tau(i)}$ as the sepset between
$C_{\tau(i)}$ and $C_{\tau(i+1)}$, i.e. the set of variables shared between these two cliques.
By the correspondence between clique trees and induced graphs, $S_{\tau(i)}$ is exactly the unmarked neighbours of $z_{\tau(i)}$ in the partially constructed induced graph at step $n+1-i$.  
Therefore, setting the parents of $z_{\tau(i)}$ to be its unmarked neighbours in Line 11 of Algorithm 1 constructs $\mathcal{H}$ with the factorization
\begin{align}
\label{eq:qfact}
q_\psi(\mathbf{z}\mid\mathbf{x})=\prod_{i=1}^{n} q_{i}(z_{\tau(i)}\mid S_{\tau(i)}),
\end{align}
which is of the form of~\eqref{eq:H-fact} with $\text{Pa}_\mathcal{H}(z_{\tau(i)})=S_{\tau(i)}$.

By construction, all $z_{\tau(>i)}=\{z_{\tau(i+1)},\ldots,z_{\tau(n)}\}$ 
are upstream in the clique tree from $z_{\tau(i)}$ (and thus downstream in the factorization), meaning they
will never be included by $S_{\tau(i)}$ .  Furthermore, the sepset property of
clique trees \citep[Theorem 10.2]{KollerFriedman2009} guarantees that
 $z_{\tau(i)}$ is independent from $z_{\tau(<i)}\setminus S_{\tau(i)}$ given $S_{\tau(i)}$.
Therefore, we have that $p(z_{\tau(i)}\mid S_{\tau(i)})=p(z_{\tau(i)}\mid z_{\tau(<i)},\mathbf{x})$ for
each variable and so
\begin{align}
p(\mathbf{z}\mid\mathbf{x})=\prod^n_{i=1}p(z_{\tau(i)}\mid z_{\tau(<i)},\mathbf{x})=\prod^n_{i=1}p(z_{\tau(i)}\mid S_{\tau(i)}).
\end{align}
This is the same as the factorization produced by NaMI, as given in~\eqref{eq:qfact}, and so we conclude
that $\mathcal{H}$ is an I-Map of $\mathcal{G}$ and thus a faithful inverse.

The minimality now follows from the fact that the sepset $S_{\tau(i)}$ is also the minimal separating set (see, e.g., the proof of \citep[Theorem 4.12]{KollerFriedman2009}).  In other words, for each $i$, there is no $T_i\subsetneq S_{\tau(i)}$ such
that $p(z_{\tau(i)}\mid T_i) = p(z_{\tau(i)}\mid S_{\tau(i)})$. 
Suppose we were to remove an edge, $z_{\tau(i)}\leftarrow y_j$, from $\mathcal{H}$, where
$y_j \in \{\mathbf{x},z_{\tau(<i)}\}$ (remembering that by construction we have no edges from $z_{\tau(>i)}$ to $z_{\tau(i)}$). This edge would have been constructed due to sepset $S_{\tau(i)}$.
If removing the edge did not make $\mathcal{H}$ unfaithful to $\mathcal{G}$, then this would imply that
$p(z_{\tau(i)}\mid S_{\tau(i)}\setminus\{y_j\}) = p(z_{\tau(i)}\mid S_{\tau(i)})$ as none of the other
factors will change.  But we have already shown this is not possible.
Hence, by contradiction, $\mathcal{H}$ is minimally faithful.
\end{proof}
%%% Local Variables:
%%% mode: latex
%%% TeX-master: "main"
%%% End: